\documentclass{article}
\usepackage[T1]{fontenc}
\usepackage{iclr2027_conference,times}

\usepackage{graphicx}
\graphicspath{{figures/}}
\usepackage{booktabs}
\usepackage{multirow}
\usepackage{array}
\usepackage{needspace}
\usepackage{placeins}
\usepackage{float}
\usepackage{amsmath,amssymb}
\usepackage{microtype}
\usepackage{xcolor}
\usepackage{hyperref}
\hypersetup{
  hidelinks,
  pdftitle={HEART-JEPA: Learning Cross-View Representations for Cardiac MRI with Joint-Embedding Predictive Architectures},
  pdfauthor={Xingzhong Zhao, Lei You, Ziqian Xie, Tian Xia, Degui Zhi}
}
\usepackage{url}

\newcommand{\method}{HEART-JEPA}
\newcommand{\Rtwo}{\ensuremath{R^{2}}}
\newcommand{\sg}{\operatorname{sg}}
\newcommand{\FormalPZeroCosineMean}{0.932}
\newcommand{\FormalPZeroCosineSD}{0.021}
\newcommand{\FormalPZeroRetrievalMean}{41.13}
\newcommand{\FormalPZeroRetrievalSD}{5.49}
\newcommand{\FormalPZeroMorphologyMean}{0.743}
\newcommand{\FormalPZeroMorphologySD}{0.007}
\newcommand{\FormalPZeroStructuralTransferMean}{0.479}
\newcommand{\FormalPZeroStructuralTransferSD}{0.022}
\newcommand{\FormalPTwentyFiveCosineMean}{0.254}
\newcommand{\FormalPTwentyFiveCosineSD}{0.037}
\newcommand{\FormalPTwentyFiveRetrievalMean}{0.207}
\newcommand{\FormalPTwentyFiveRetrievalSD}{0.190}
\newcommand{\FormalPTwentyFiveMorphologyMean}{0.688}
\newcommand{\FormalPTwentyFiveMorphologySD}{0.016}
\newcommand{\FormalPTwentyFiveStructuralTransferMean}{0.062}
\newcommand{\FormalPTwentyFiveStructuralTransferSD}{0.104}
\newcommand{\FormalPFiftyCosineMean}{0.114}
\newcommand{\FormalPFiftyCosineSD}{0.025}
\newcommand{\FormalPFiftyRetrievalMean}{0.0109}
\newcommand{\FormalPFiftyRetrievalSD}{0.0042}
\newcommand{\FormalPFiftyMorphologyMean}{0.618}
\newcommand{\FormalPFiftyMorphologySD}{0.054}
\newcommand{\FormalPFiftyStructuralTransferMean}{-0.965}
\newcommand{\FormalPFiftyStructuralTransferSD}{0.524}
\newcommand{\FormalPSeventyFiveCosineMean}{0.119}
\newcommand{\FormalPSeventyFiveCosineSD}{0.007}
\newcommand{\FormalPSeventyFiveRetrievalMean}{0.0146}
\newcommand{\FormalPSeventyFiveRetrievalSD}{0.0039}
\newcommand{\FormalPSeventyFiveMorphologyMean}{0.660}
\newcommand{\FormalPSeventyFiveMorphologySD}{0.011}
\newcommand{\FormalPSeventyFiveStructuralTransferMean}{-1.327}
\newcommand{\FormalPSeventyFiveStructuralTransferSD}{0.235}
\newcommand{\FormalPOneCosineMean}{0.140}
\newcommand{\FormalPOneCosineSD}{0.069}
\newcommand{\FormalPOneRetrievalMean}{0.0277}
\newcommand{\FormalPOneRetrievalSD}{0.0101}
\newcommand{\FormalPOneMorphologyMean}{0.677}
\newcommand{\FormalPOneMorphologySD}{0.021}
\newcommand{\FormalPOneStructuralTransferMean}{-1.039}
\newcommand{\FormalPOneStructuralTransferSD}{0.393}
\newcommand{\StrictPZeroCenteredCosineMean}{0.787}
\newcommand{\StrictPZeroCenteredCosineSD}{0.027}
\newcommand{\StrictPZeroCKAMean}{0.850}
\newcommand{\StrictPZeroCKASD}{0.007}
\newcommand{\StrictPZeroRetrievalMean}{41.24}
\newcommand{\StrictPZeroRetrievalSD}{5.43}
\newcommand{\StrictPZeroViewIdentityMean}{99.98}
\newcommand{\StrictPZeroViewIdentitySD}{0.0024}
\newcommand{\StrictPZeroMorphologyMean}{0.747}
\newcommand{\StrictPZeroMorphologySD}{0.007}
\newcommand{\StrictPZeroStructuralTransferMean}{0.477}
\newcommand{\StrictPZeroStructuralTransferSD}{0.020}
\newcommand{\StrictPOneCenteredCosineMean}{0.046}
\newcommand{\StrictPOneCenteredCosineSD}{0.021}
\newcommand{\StrictPOneCKAMean}{0.055}
\newcommand{\StrictPOneCKASD}{0.030}
\newcommand{\StrictPOneRetrievalMean}{0.0279}
\newcommand{\StrictPOneRetrievalSD}{0.0103}
\newcommand{\StrictPOneViewIdentityMean}{99.98}
\newcommand{\StrictPOneViewIdentitySD}{0.0175}
\newcommand{\StrictPOneMorphologyMean}{0.680}
\newcommand{\StrictPOneMorphologySD}{0.021}
\newcommand{\StrictPOneStructuralTransferMean}{-1.101}
\newcommand{\StrictPOneStructuralTransferSD}{0.393}
\newcommand{\PhasePZeroCircularMAEMean}{0.061}
\newcommand{\PhasePZeroCircularMAESD}{0.011}
\newcommand{\PhasePZeroTopOneMean}{68.49}
\newcommand{\PhasePZeroTopOneSD}{3.56}
\newcommand{\PhasePZeroCosineMean}{0.757}
\newcommand{\PhasePZeroCosineSD}{0.030}
\newcommand{\PhasePZeroMarginMean}{0.150}
\newcommand{\PhasePZeroMarginSD}{0.034}
\newcommand{\PhasePOneCircularMAEMean}{0.196}
\newcommand{\PhasePOneCircularMAESD}{0.008}
\newcommand{\PhasePOneTopOneMean}{25.41}
\newcommand{\PhasePOneTopOneSD}{2.55}
\newcommand{\PhasePOneCosineMean}{0.086}
\newcommand{\PhasePOneCosineSD}{0.009}
\newcommand{\PhasePOneMarginMean}{0.057}
\newcommand{\PhasePOneMarginSD}{0.013}
\newcommand{\ObjectivePZeroCrossNMSEMean}{0.298}
\newcommand{\ObjectivePZeroCrossNMSESD}{0.025}
\newcommand{\ObjectivePZeroSameNMSEMean}{0.270}
\newcommand{\ObjectivePZeroSameNMSESD}{0.014}
\newcommand{\ObjectivePTwentyFiveCrossNMSEMean}{0.677}
\newcommand{\ObjectivePTwentyFiveCrossNMSESD}{0.014}
\newcommand{\ObjectivePTwentyFiveSameNMSEMean}{0.371}
\newcommand{\ObjectivePTwentyFiveSameNMSESD}{0.023}
\newcommand{\ObjectivePOneCrossNMSEMean}{1.187}
\newcommand{\ObjectivePOneCrossNMSESD}{0.039}
\newcommand{\ObjectivePOneSameNMSEMean}{0.552}
\newcommand{\ObjectivePOneSameNMSESD}{0.025}
\newcommand{\ObjectivePZeroEncoderGradientCosineMean}{0.469}
\newcommand{\ObjectivePZeroEncoderGradientCosineSD}{0.078}
\newcommand{\ObjectivePZeroPredictorGradientCosineMean}{0.350}
\newcommand{\ObjectivePZeroPredictorGradientCosineSD}{0.048}
\newcommand{\ObjectivePTwentyFiveEncoderGradientCosineMean}{0.026}
\newcommand{\ObjectivePTwentyFiveEncoderGradientCosineSD}{0.005}
\newcommand{\ObjectivePTwentyFivePredictorGradientCosineMean}{0.042}
\newcommand{\ObjectivePTwentyFivePredictorGradientCosineSD}{0.014}
\newcommand{\ObjectivePOneEncoderGradientCosineMean}{0.073}
\newcommand{\ObjectivePOneEncoderGradientCosineSD}{0.024}
\newcommand{\ObjectivePOnePredictorGradientCosineMean}{0.068}
\newcommand{\ObjectivePOnePredictorGradientCosineSD}{0.037}
\newcommand{\ContrastiveCosineMean}{0.848}
\newcommand{\ContrastiveCosineSD}{0.007}
\newcommand{\ContrastiveMarginMean}{0.532}
\newcommand{\ContrastiveMarginSD}{0.001}
\newcommand{\ContrastiveRetrievalMean}{45.67}
\newcommand{\ContrastiveRetrievalSD}{1.27}
\newcommand{\ContrastiveViewIdentityMean}{99.98}
\newcommand{\ContrastiveViewIdentitySD}{0.0039}
\newcommand{\ContrastiveMorphologyMean}{0.725}
\newcommand{\ContrastiveMorphologySD}{0.002}
\newcommand{\ContrastiveStructuralTransferMean}{0.442}
\newcommand{\ContrastiveStructuralTransferSD}{0.023}
\newcommand{\ContrastiveFunctionalTransferMean}{0.351}
\newcommand{\ContrastiveFunctionalTransferSD}{0.072}
\newcommand{\HeartObjectiveCosineMean}{0.932}
\newcommand{\HeartObjectiveCosineSD}{0.021}
\newcommand{\HeartObjectiveMarginMean}{0.272}
\newcommand{\HeartObjectiveMarginSD}{0.118}
\newcommand{\HeartObjectiveRetrievalMean}{41.24}
\newcommand{\HeartObjectiveRetrievalSD}{5.43}
\newcommand{\HeartObjectiveViewIdentityMean}{99.98}
\newcommand{\HeartObjectiveViewIdentitySD}{0.0019}
\newcommand{\HeartObjectiveMorphologyMean}{0.743}
\newcommand{\HeartObjectiveMorphologySD}{0.007}
\newcommand{\HeartObjectiveStructuralTransferMean}{0.479}
\newcommand{\HeartObjectiveStructuralTransferSD}{0.022}
\newcommand{\HeartObjectiveFunctionalTransferMean}{0.447}
\newcommand{\HeartObjectiveFunctionalTransferSD}{0.023}

\newcommand{\TargetControlHeartCosineMean}{0.932}
\newcommand{\TargetControlHeartCosineSD}{0.021}
\newcommand{\TargetControlHeartRetrievalMean}{41.24}
\newcommand{\TargetControlHeartRetrievalSD}{5.43}
\newcommand{\TargetControlHeartMorphologyMean}{0.743}
\newcommand{\TargetControlHeartMorphologySD}{0.007}
\newcommand{\TargetControlHeartStructuralTransferMean}{0.479}
\newcommand{\TargetControlHeartStructuralTransferSD}{0.022}
\newcommand{\TargetControlHeartFunctionalTransferMean}{0.447}
\newcommand{\TargetControlHeartFunctionalTransferSD}{0.023}
\newcommand{\TargetControlIndependentCosineMean}{0.923}
\newcommand{\TargetControlIndependentCosineSD}{0.014}
\newcommand{\TargetControlIndependentRetrievalMean}{46.24}
\newcommand{\TargetControlIndependentRetrievalSD}{8.81}
\newcommand{\TargetControlIndependentMorphologyMean}{0.739}
\newcommand{\TargetControlIndependentMorphologySD}{0.011}
\newcommand{\TargetControlIndependentStructuralTransferMean}{0.470}
\newcommand{\TargetControlIndependentStructuralTransferSD}{0.027}
\newcommand{\TargetControlIndependentFunctionalTransferMean}{0.487}
\newcommand{\TargetControlIndependentFunctionalTransferSD}{0.020}
\newcommand{\TargetControlGlobalCosineMean}{0.988}
\newcommand{\TargetControlGlobalCosineSD}{0.003}
\newcommand{\TargetControlGlobalMarginMean}{0.008}
\newcommand{\TargetControlGlobalMarginSD}{0.002}
\newcommand{\TargetControlGlobalRetrievalMean}{28.18}
\newcommand{\TargetControlGlobalRetrievalSD}{8.07}
\newcommand{\TargetControlGlobalMorphologyMean}{0.775}
\newcommand{\TargetControlGlobalMorphologySD}{0.002}
\newcommand{\TargetControlGlobalStructuralTransferMean}{-0.550}
\newcommand{\TargetControlGlobalStructuralTransferSD}{0.130}
\newcommand{\TargetControlGlobalFunctionalTransferMean}{-0.096}
\newcommand{\TargetControlGlobalFunctionalTransferSD}{0.251}

\newcommand{\TraitHeartTwoFourOneOneOneTransferMean}{0.569}
\newcommand{\TraitHeartTwoFourOneOneOneTransferSD}{0.042}
\newcommand{\TraitHeartTwoFourOneOneFourTransferMean}{0.112}
\newcommand{\TraitHeartTwoFourOneOneFourTransferSD}{0.042}
\newcommand{\TraitHeartTwoFourOneZeroFiveTransferMean}{0.596}
\newcommand{\TraitHeartTwoFourOneZeroFiveTransferSD}{0.021}
\newcommand{\TraitHeartTwoFourOneFourZeroTransferMean}{0.609}
\newcommand{\TraitHeartTwoFourOneFourZeroTransferSD}{0.022}
\newcommand{\TraitHeartRAMaxFourToThreeMean}{-0.284}
\newcommand{\TraitHeartRAMaxFourToThreeSD}{0.192}
\newcommand{\TraitHeartRAMinFourToThreeMean}{-0.366}
\newcommand{\TraitHeartRAMinFourToThreeSD}{0.267}

\newcommand{\EFHeartCheckpoints}{3}
\newcommand{\EFHeartLVEFMAEMean}{3.1888}
\newcommand{\EFHeartLVEFMAESD}{0.0022}
\newcommand{\EFHeartLVEFRMean}{0.7563}
\newcommand{\EFHeartLVEFRSD}{0.0005}
\newcommand{\EFHeartRVEFMAEMean}{3.3211}
\newcommand{\EFHeartRVEFMAESD}{0.0017}
\newcommand{\EFHeartRVEFRMean}{0.7092}
\newcommand{\EFHeartRVEFRSD}{0.0004}
\newcommand{\EFGlobalCheckpoints}{3}
\newcommand{\EFGlobalLVEFMAEMean}{3.2016}
\newcommand{\EFGlobalLVEFMAESD}{0.0014}
\newcommand{\EFGlobalLVEFRMean}{0.7527}
\newcommand{\EFGlobalLVEFRSD}{0.0003}
\newcommand{\EFGlobalRVEFMAEMean}{3.2307}
\newcommand{\EFGlobalRVEFMAESD}{0.0023}
\newcommand{\EFGlobalRVEFRMean}{0.7267}
\newcommand{\EFGlobalRVEFRSD}{0.0003}
\newcommand{\EFSameViewCheckpoints}{3}
\newcommand{\EFSameViewLVEFMAEMean}{3.4223}
\newcommand{\EFSameViewLVEFMAESD}{0.0018}
\newcommand{\EFSameViewLVEFRMean}{0.7085}
\newcommand{\EFSameViewLVEFRSD}{0.0006}
\newcommand{\EFSameViewRVEFMAEMean}{3.5126}
\newcommand{\EFSameViewRVEFMAESD}{0.0008}
\newcommand{\EFSameViewRVEFRMean}{0.6689}
\newcommand{\EFSameViewRVEFRSD}{0.0003}
\newcommand{\EFContrastiveCheckpoints}{3}
\newcommand{\EFContrastiveLVEFMAEMean}{3.4180}
\newcommand{\EFContrastiveLVEFMAESD}{0.0017}
\newcommand{\EFContrastiveLVEFRMean}{0.7084}
\newcommand{\EFContrastiveLVEFRSD}{0.0005}
\newcommand{\EFContrastiveRVEFMAEMean}{3.4674}
\newcommand{\EFContrastiveRVEFMAESD}{0.0035}
\newcommand{\EFContrastiveRVEFRMean}{0.6791}
\newcommand{\EFContrastiveRVEFRSD}{0.0008}
\newcommand{\EFCMRTransformerCheckpoints}{1}
\newcommand{\EFCMRTransformerLVEFMAEMean}{3.2390}
\newcommand{\EFCMRTransformerLVEFMAESD}{0.0029}
\newcommand{\EFCMRTransformerLVEFRMean}{0.7451}
\newcommand{\EFCMRTransformerLVEFRSD}{0.0004}
\newcommand{\EFCMRTransformerRVEFMAEMean}{3.5207}
\newcommand{\EFCMRTransformerRVEFMAESD}{0.0029}
\newcommand{\EFCMRTransformerRVEFRMean}{0.6696}
\newcommand{\EFCMRTransformerRVEFRSD}{0.0010}
\newcommand{\EFCineMACheckpoints}{1}
\newcommand{\EFCineMALVEFMAEMean}{4.0860}
\newcommand{\EFCineMALVEFMAESD}{0.0031}
\newcommand{\EFCineMALVEFRMean}{0.5254}
\newcommand{\EFCineMALVEFRSD}{0.0011}
\newcommand{\EFCineMARVEFMAEMean}{3.9496}
\newcommand{\EFCineMARVEFMAESD}{0.0043}
\newcommand{\EFCineMARVEFRMean}{0.5427}
\newcommand{\EFCineMARVEFRSD}{0.0014}
\newcommand{\EFViTaMAECheckpoints}{1}
\newcommand{\EFViTaMAELVEFMAEMean}{3.8391}
\newcommand{\EFViTaMAELVEFMAESD}{0.0124}
\newcommand{\EFViTaMAELVEFRMean}{0.6043}
\newcommand{\EFViTaMAELVEFRSD}{0.0029}
\newcommand{\EFViTaMAERVEFMAEMean}{3.9735}
\newcommand{\EFViTaMAERVEFMAESD}{0.0050}
\newcommand{\EFViTaMAERVEFRMean}{0.5530}
\newcommand{\EFViTaMAERVEFRSD}{0.0017}
\newcommand{\EFLVEFReductionVsContrastive}{6.7}
\newcommand{\EFRVEFReductionVsContrastive}{4.2}

\title{\method{}: Learning Cross-View Representations for Cardiac MRI with Joint-Embedding Predictive Architectures
}
\iclrfinalcopy
\author{%
  Xingzhong Zhao\textsuperscript{1}, Lei You\textsuperscript{2},
  Ziqian Xie\textsuperscript{1}, Tian Xia\textsuperscript{1},
  Degui Zhi\textsuperscript{1}\\[5pt]
  \normalfont\textsuperscript{1}Yale University\\
  \normalfont\textsuperscript{2}McWilliams School of Biomedical Informatics, UTHealth Houston
}

\begin{document}
\maketitle
\fancyhead[L]{\small Preprint}
\begin{abstract}
Cine cardiac magnetic resonance imaging (CMR) captures the same heart from complementary anatomical views. We ask whether a predictor trained on one view can be reused on another. We introduce HEART-JEPA, a joint-embedding predictive architecture that learns from paired views by predicting masked latent features across anatomical planes without spatially matched patches. We pretrained HEART-JEPA on 39,992 UK Biobank participants. On 15,935 participants excluded from pretraining, structural phenotype prediction achieved a cross-view $R^2$ of 0.48 when both the source standardizer and predictor were held fixed. In matched experiments, cross-view pretraining improved direct transfer over same-view pretraining. A target-specific probe reached $R^2=0.536$ on the same-view features. Target-moment normalization and paired Procrustes alignment both left direct transfer below zero for that model. In the expanded cohort, HEART-JEPA also outperformed cross-view contrastive learning on cardiac phenotype probing and on structural and functional transfer. In a separate seven-model benchmark, linear probes on concatenated frozen features from all three views achieved the lowest mean absolute error for left ventricular ejection fraction (3.19 percentage points), with competitive results for right ventricular ejection fraction.
\end{abstract}

\section{Introduction}

Cardiac magnetic resonance imaging (CMR) is used to assess cardiac anatomy, function, and tissue characteristics noninvasively \citep{kramer2020cmrprotocols}. Cine CMR records motion throughout the cardiac cycle and supports measurements such as ventricular volume, myocardial mass, and ejection fraction. Clinicians acquire these images in several anatomical planes, each showing different parts of the same heart. Automated measurements often rely on expert annotations and task-specific training \citep{fu2025cinema,shad2026cmr}. A shared representation gives these tasks a common starting point.

Several approaches learn CMR representations from multiple views or from accompanying clinical data. CineMA encodes multiple cardiac views and reconstructs masked image patches \citep{fu2025cinema}. ViTa pairs multiview reconstruction with image--tabular alignment \citep{zhang2025vita}, and CMR Transformer learns from paired cine videos and clinical reports \citep{shad2026cmr}. MR-JEPA extends LeJEPA \citep{balestriero2025lejepa} to spatiotemporal CMR and combines anatomical views through gated attention for downstream prediction \citep{jacob2026mrjepa}.

The two-, three-, and four-chamber long-axis views share aspects of cardiac morphology and motion, although they cover different anatomy. The same image coordinates can therefore refer to different structures. In HEART-JEPA, we use features from one view to predict masked latent features in another. The inputs are paired by participant, and the objective requires neither patch correspondences nor physiological frame alignment.

At evaluation, we fit a cardiac phenotype predictor to one view and apply it to another with the same standardizer and coefficients. We use \emph{predictive compatibility} for this ability to reuse a fitted predictor across views. We pretrain on 39,992 UK Biobank participants and compare same-view and cross-view targets under matched training conditions. Cross-view pretraining improves direct transfer on 15,935 participants excluded from pretraining. A separate target-specific fit tests how well each representation supports phenotype prediction when the probe is trained on the target view.

\noindent\textbf{Contributions.}
\textbf{(i)} We train HEART-JEPA to predict masked latent features across paired long-axis views on a fixed video-JEPA backbone.
\textbf{(ii)} We measure direct transfer with a fixed phenotype predictor and test the effects of target normalization, paired alignment, and refitting separately.
\textbf{(iii)} Matched comparisons show that the pretraining objectives rank differently on probing, retrieval, and direct transfer, and that performance varies non-monotonically as we mix same-view and cross-view targets.

\section{Related Work}

\paragraph{Cross-view and multimodal prediction.}
I-JEPA and V-JEPA predict latent content \citep{assran2023ijepa,bardes2024vjepa}, whereas MAE and VideoMAE reconstruct pixels \citep{he2022mae,tong2022videomae}. CroCo uses cross-view pixel completion to learn geometric correspondence \citep{weinzaepfel2022croco}. M3-JEPA aligns modalities with a Multi-Gate Mixture-of-Experts predictor, contrastive learning, and regularization \citep{lei2025m3jepa}; CrossJEPA predicts rendered image features to learn point-cloud representations \citep{perera2025crossjepa}. We build on cross-observation prediction, using a shared encoder and exponential-moving-average (EMA) target branch for anatomical views within one modality. Cross-view prediction can be solved by a predictor without making encoder coordinates compatible. We therefore evaluate frozen-encoder readout reuse separately from latent prediction and similarity, without claiming matched empirical superiority over M3-JEPA.

\paragraph{Cardiac representation learning.}
CineMA uses multi-view masked reconstruction, ViTa combines images with tabular information, and CMR Transformer uses image--report supervision \citep{fu2025cinema,zhang2025vita,shad2026cmr}. MR-JEPA extends LeJEPA-style self-supervision to multi-sequence CMR and uses downstream gated attention to combine view features \citep{jacob2026mrjepa}. HEART-JEPA instead makes anatomical view relation part of the pretraining target and evaluates unchanged source-to-target readout reuse.

\section{Cross-View Latent Prediction}
\label{sec:method}

\subsection{Predictive compatibility}

Let $z_n^v$ be the frozen EMA encoder representation of participant $n$ in view $v$. We fit a feature standardizer $S_c$ and phenotype prediction model $f_c$ on source-view training features, then evaluate a different target view $t$:

\begin{equation}
\hat{y}_{n,c\rightarrow t}
=
f_c\!\left(S_c\!\left(z_n^t\right)\right),
\qquad c \neq t .
\label{eq:cross_view_transfer}
\end{equation}

Training and testing use different participants. Both $S_c$ and $f_c$, including the downstream coefficients and intercept, remain fixed; the view-conditioned pretraining predictor is excluded. Thus, performance measures the frozen encoder representations rather than a learned translation applied at test time. Direct transfer allows neither target-derived feature normalization nor output calibration. Target-specific refitting instead estimates a new prediction model from target-view features and labels.

Transferred $R^2$ combines feature-moment, predictive-direction, and anatomical-coverage effects; a negative value does not imply absence of phenotype information. The 2Ch/3Ch views predominantly capture the left heart, whereas 4Ch includes both sides \citep{kramer2020cmrprotocols}. Right-heart prediction from left-heart views can exploit indirect correlates; Appendix~\ref{app:idp} reports left- and right-heart results separately.

\subsection{Masked cross-view prediction}
\label{sec:architecture}

Each paired long-axis video contains 50 frames at $128\times128$ resolution. For input video $x^v$, temporal tube size 5 and spatial patch size 16 yield the lattice $\mathcal G=\{1,\ldots,10\}\times\{1,\ldots,8\}\times\{1,\ldots,8\}$. For context--target pair $(v_c,v_t)$, positional embeddings are added before masking. The online encoder processes visible context tokens, $h^c=E_\theta(x^{v_c};\mathcal G\setminus M_c)$; the EMA encoder processes the full target video, $H^t=E_\phi(x^{v_t};\mathcal G)$. With target queries $Q_t$ and view-conditioned predictor $P_\psi$, the loss is

\begin{equation}
\mathcal L_{\mathrm{token}}=
\frac{1}{|Q_t|}\sum_{q\in Q_t}
\left\|P_\psi(h^c,q,v_t)-\sg(H_q^t)\right\|_2^2.
\label{eq:token}
\end{equation}

For each $q$, the predictor receives visible context features, the target-position embedding, and target-view identity. Stop-gradient $\sg$ blocks target-branch backpropagation, while $\phi\leftarrow\tau\phi+(1-\tau)\theta$ with $\tau=0.996$. The encoder and predictor have 12 and six Transformer blocks, respectively.

HEART-JEPA uses $v_c\ne v_t$ and, by default, $Q_t=M_c$. Shared indices refer to separately normalized grids, not registered anatomy. Training requires paired videos but no cross-view image registration, supplied patch matching, or physiological frame alignment. Independently sampled queries test this index coupling (Section~\ref{sec:target_relation_controls}). Under the fixed-target assumptions in Appendix~\ref{app:derivation}, optimal prediction averages out context-unpredictable target variation; encoder compatibility remains an empirical question. Appendix~\ref{app:implementation} gives implementation details.

\begin{figure}[t]
\centering
\includegraphics[width=\linewidth]{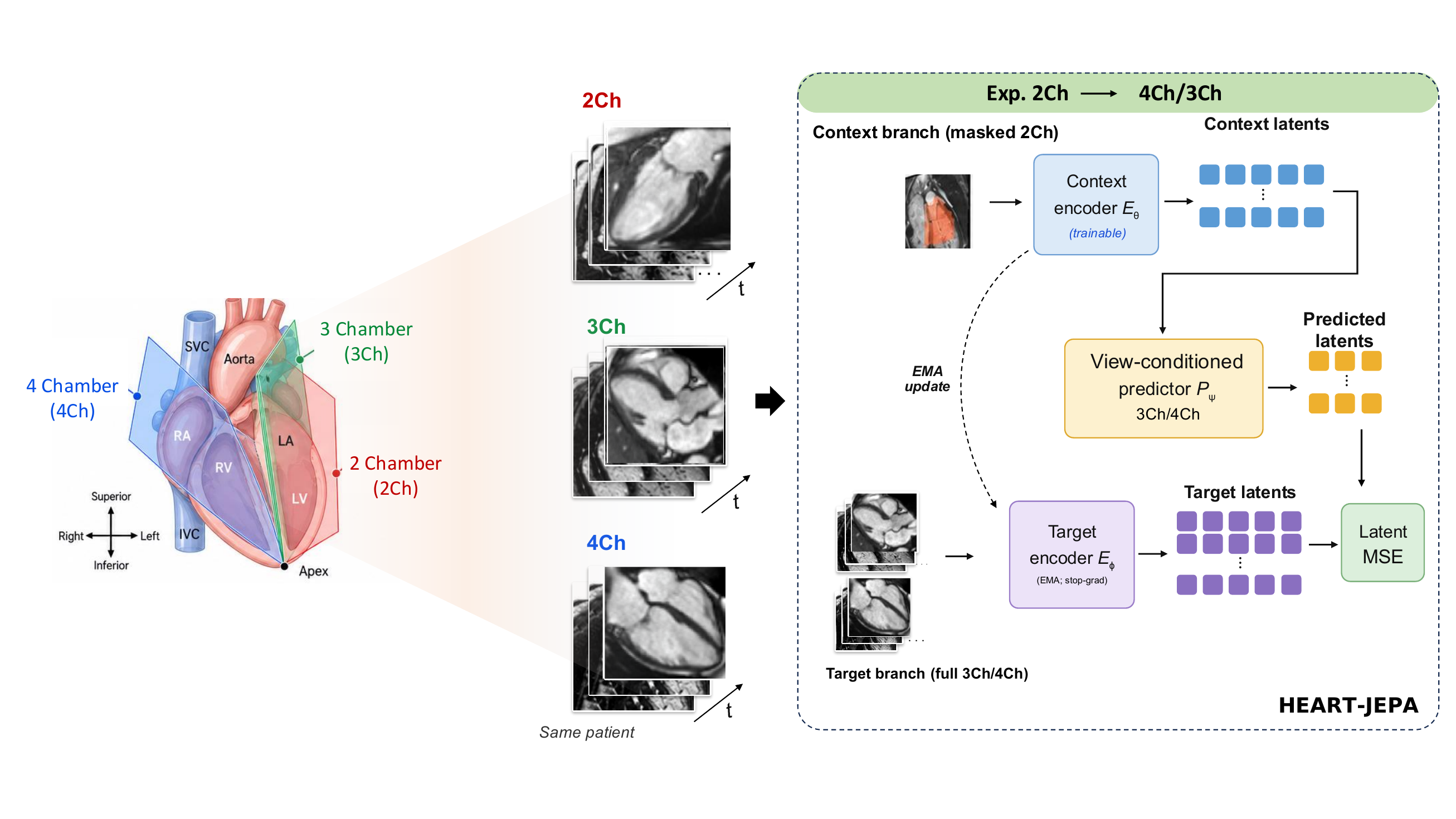}
\caption{\textbf{Overview of HEART-JEPA.}
Paired 2-, 3-, and 4-chamber cine CMR views from the same subject provide complementary observations of the same heart. The right panel illustrates a 2Ch$\rightarrow$3Ch/4Ch training configuration: a masked 2Ch context view is encoded by the trainable context encoder $E_{\theta}$, and a view-conditioned predictor $P_{\psi}$ predicts latent features at queried positions in the full 3Ch or 4Ch target view. Target features are produced by the EMA-updated encoder $E_{\phi}$ with stop-gradient, and training minimizes the latent-space mean-squared error between predicted and target features. }
\label{fig:method}
\end{figure}

\section{Experimental Design}
\label{sec:experiments}

\paragraph{Data and preprocessing.}
Pretraining uses 39{,}992 UK Biobank participants with all three long-axis views: 37{,}993 for optimization and 1{,}999 for monitoring \citep{bycroft2018ukb}. This common one-video-per-view structure isolates target relation without changing input dimensionality; short-axis stacks have different geometry and are outside the study. Controlled models share segmentation-derived bounding-box crops, per-frame z-scoring, and resizing. Masks supply the crop prior, not encoder inputs. Videos retain the first 50 frames or repeat the last when shorter, without temporal interpolation or physiological registration.

\paragraph{Cohorts.}
C2 contains 15{,}935 participants absent from pretraining; C3 and C4 are fixed subsets (Table~\ref{tab:cohorts}). The full mixture sweep and objective controls also use a separate expanded set of 17{,}841 participants (Appendix~\ref{app:audit}).

\begin{table}[t]
\centering
\caption{\textbf{Study cohorts and evaluation roles.} The expanded analysis set is described in Appendix~\ref{app:audit}.}
\label{tab:cohorts}
\small
\setlength{\tabcolsep}{4pt}
\begin{tabular}{cclp{0.38\linewidth}}
\toprule
\textbf{Label} &
\textbf{Sample size} &
\textbf{Definition} &
\textbf{Evaluation role} \\
\midrule

C1 &
39,992 &
Pretraining cohort &
Self-supervised pretraining and monitoring \\

C2 &
15,935 &
Participant-disjoint from C1 &
Cardiac probing, direct transfer, EF, and disease evaluation \\

C3 &
4,096 &
Fixed subset of C2 &
Centered representation geometry and linear CKA \\

C4 &
512 &
Fixed subset of C2 &
Temporal-token matching across views \\

C5 &
360 &
External M\&Ms-2 cohort; 4Ch only &
External cardiac classification \\
\bottomrule
\end{tabular}
\end{table}

\paragraph{Controlled training.}
We vary $p_{\mathrm{same}}\in\{0,0.25,0.5,0.75,1\}$ within three matched seeds, fixing architecture, MultiBlockMask3D masking, preprocessing, ordered splits, and optimization. AdamW uses batch size 48 for 276,850 updates; evaluation uses the epoch-350 checkpoint without early stopping. Additional controls use independent queries, global non-contrastive normalized regression (Global HEART-JEPA), or cross-view EMA-InfoNCE. These objective-family controls change multiple head/loss factors, so they do not isolate target granularity alone (Appendices~\ref{app:implementation}--\ref{app:audit}).

\paragraph{Cardiac probing and transfer.}
The 17 imaging-derived phenotypes (IDPs) comprise 13 structural and four functional traits (Appendix~\ref{app:idp}). Fitted probes use standardized linear models on concatenated 2Ch/3Ch/4Ch features and five-fold cross-validated $R^2$. Direct transfer uses a source-trained standardized ridge ($\alpha=1$) across six ordered view pairs and the same five participant folds. Scalers and prediction models are fitted only on training participants; all conditions reuse the existing folds and frozen features. On C2, we additionally evaluate direct Pearson $r$, training-only target-feature moments, paired orthogonal Procrustes, and target-specific refitting. Only the last fits a new supervised target readout; corrections leave the source ridge unchanged. Primary diagnostic scores average 13 traits, six directions, and five folds within seed, then report mean $\pm$ sample SD over three pretrained seeds (Appendix~\ref{app:compatibility}).

\paragraph{Ventricular function.}
Left- and right-ventricular ejection fraction (LVEF/RVEF) form a separate seven-model benchmark using the same C2 participants, concatenated views, probe specification, and outer folds. These clinically familiar endpoints complement the broader IDP panel rather than replace its structural and functional traits. MAE is in EF percentage points; $R$ denotes Pearson correlation. Appendix~\ref{app:ef_protocol} specifies the repeated-CV aggregation.

\paragraph{Other evaluation roles.}
Cosine participant retrieval, centered cosine and linear centered-kernel alignment (CKA) on C3, and ancillary sequence-position matching on C4 characterize representation organization. Retrieval uses no learned metric and evaluates C2 queries against the fixed 17{,}841-participant gallery. Conventional utility is assessed with disease-label probes, available-view/limited-label IDP probes, and external M\&Ms-2 classification \citep{mnms2challenge}. These allow task-specific fitting, unlike direct transfer. External comparisons use available frozen features and their specified adapters. All evaluation splits are participant-level.

\section{Results}
\label{sec:shared_results}

\subsection{Cross-view targets improve direct transfer}
\label{sec:targetmix}
\label{sec:strict}

\begin{figure}[t]
\centering
\includegraphics[width=\linewidth]{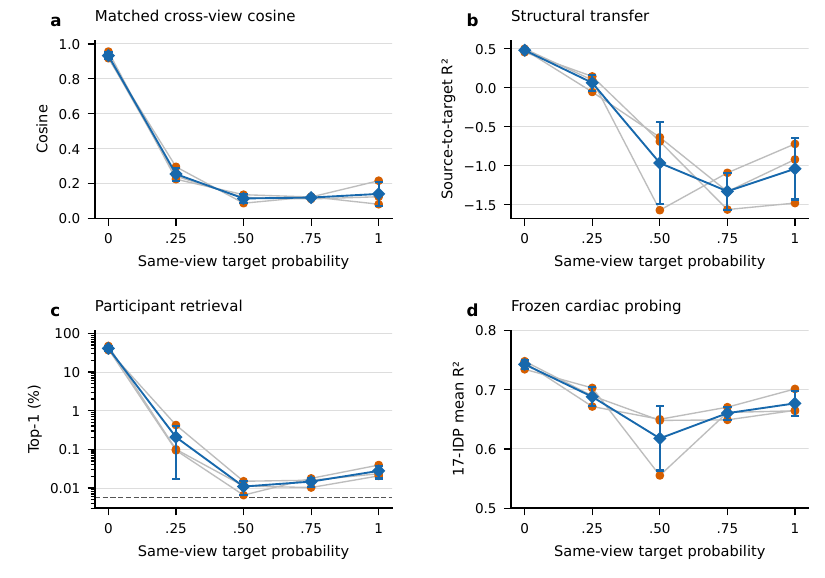}
\caption{\textbf{Effect of target source on learned representations.}
(a) Matched cross-view cosine similarity,
(b) structural transfer without target-view refitting,
(c) participant retrieval, and
(d) fitted three-view cardiac probing.
Points denote individual training seeds, gray lines connect matched seeds, and
diamonds show mean $\pm$ sample SD. Retrieval is shown on a logarithmic
percentage scale, with the dashed line indicating random top-1 performance.
All conditions are trained from scratch under the same training budget.}
\label{fig:targetmix}
\end{figure}

On the expanded analysis set, structural-transfer $R^2$ rises from
$\FormalPOneStructuralTransferMean\pm\FormalPOneStructuralTransferSD$ for pure same-view pretraining to
$\FormalPZeroStructuralTransferMean\pm\FormalPZeroStructuralTransferSD$ for pure cross-view pretraining. Fitted three-view probing also improves, from
$\FormalPOneMorphologyMean\pm\FormalPOneMorphologySD$ to
$\FormalPZeroMorphologyMean\pm\FormalPZeroMorphologySD$
(Figure~\ref{fig:targetmix}).

Excluding every pretraining participant preserves the large transfer difference. On C2, HEART-JEPA achieves
$\StrictPZeroStructuralTransferMean\pm\StrictPZeroStructuralTransferSD$ versus
$\StrictPOneStructuralTransferMean\pm\StrictPOneStructuralTransferSD$ for same-view JEPA.
Retrieval and centered geometry also improve (Table~\ref{tab:strict_endpoints}; Figure~\ref{fig:formal_diagnostics}b--c), while view classification exceeds 99.97\% across mixture conditions. Thus, transferable features retain view identity. Transfer is stronger for left-ventricular wall thickness, myocardial mass, and left-atrial volumes than for right-atrial volumes (Appendix~\ref{app:idp}).

\subsection{Readout reuse and target-view information}
\label{sec:compatibility}

The C2 diagnostics reproduce the original direct-transfer endpoints (Table~\ref{tab:compatibility}). Direct Pearson $r$ is $0.742\pm0.010$ for HEART-JEPA versus $0.298\pm0.021$ for same-view JEPA, so a global output offset or positive scaling cannot explain the weaker association. Target-moment normalization improves HEART-JEPA transfer but does not recover same-view transfer; neither does paired Procrustes, despite reducing training-feature discrepancy in all fitted maps.

Target refitting uses labeled target-feature examples and is an information-availability control, not an unlabeled correction or strict upper bound. It remains useful for both models ($R^2=0.654\pm0.010$ versus $0.536\pm0.025$). HEART-JEPA therefore combines stronger linearly accessible target information with better unchanged readout reuse; compatibility alone cannot explain the full gain. Contrastive learning is competitive. These C2 results are separate from the expanded-set controls below. Appendix~\ref{app:compatibility} reports the protocol and directional results.

\begin{table}[t]
\centering
\caption{\textbf{Structural predictive-compatibility diagnostics on C2.}
Mean $\pm$ sample SD across three pretrained seeds; 13 traits, six directions,
and five folds are averaged within seed. Only target refitting uses target training
labels. Moments and Procrustes use target training features; no condition fits
on held-out features or labels. Direct $r$ uses the direct-transfer predictions.}
\label{tab:compatibility}
\footnotesize
\setlength{\tabcolsep}{2.5pt}
\begin{tabular}{@{}lccccc@{}}
\toprule
Model & Direct $R^2$ & Direct $r$ & Moments $R^2$ & Procrustes $R^2$ & Target refit $R^2$ \\
\midrule
HEART-JEPA & $0.477\pm0.020$ & $0.742\pm0.010$ & $0.538\pm0.016$ & $0.461\pm0.048$ & $0.654\pm0.010$ \\
Same-view JEPA & $-1.101\pm0.393$ & $0.298\pm0.021$ & $-1.216\pm0.344$ & $-2.606\pm1.010$ & $0.536\pm0.025$ \\
Cross-view contrastive & $0.435\pm0.021$ & $0.723\pm0.007$ & $0.497\pm0.007$ & $0.377\pm0.011$ & $0.640\pm0.003$ \\
\bottomrule
\end{tabular}
\end{table}

\subsection{Probing, retrieval, and transfer rank objectives differently}
\label{sec:target_relation_controls}
\label{sec:contrastive}

On the expanded analysis set, Global HEART-JEPA has higher fitted-probe $R^2$ than HEART-JEPA
($\TargetControlGlobalMorphologyMean\pm\TargetControlGlobalMorphologySD$ versus
$\HeartObjectiveMorphologyMean\pm\HeartObjectiveMorphologySD$), yet negative direct transfer
($\TargetControlGlobalStructuralTransferMean\pm\TargetControlGlobalStructuralTransferSD$;
Figure~\ref{fig:formal_diagnostics}a).

Contrastive learning has higher mean C2-query retrieval
($\ContrastiveRetrievalMean\pm\ContrastiveRetrievalSD\%$ versus
$\HeartObjectiveRetrievalMean\pm\HeartObjectiveRetrievalSD\%$), but HEART-JEPA has higher fitted probing and structural/functional transfer in all three paired seeds. Structural transfer is
$\HeartObjectiveStructuralTransferMean\pm\HeartObjectiveStructuralTransferSD$ versus
$\ContrastiveStructuralTransferMean\pm\ContrastiveStructuralTransferSD$; contrastive learning has the larger cosine margin in all three seeds and higher retrieval in two (Appendix~\ref{app:contrastive}). The ranking therefore depends on whether the evaluation fits a new readout, retrieves a participant, or reuses an existing readout.

Independent target queries retain structural transfer
($\TargetControlIndependentStructuralTransferMean\pm\TargetControlIndependentStructuralTransferSD$ versus
$\TargetControlHeartStructuralTransferMean\pm\TargetControlHeartStructuralTransferSD$ for shared queries).
This removes exact index coupling while preserving incidental overlap (Jaccard $0.682\pm0.072$); Appendix~\ref{app:target_relation_controls} gives the full comparison.

\subsection{Target mixtures have non-monotonic effects}
\label{sec:mixture}

The matched target-source sweep tests whether the two relations interpolate, rather than tuning for the best mixture. Adding 25\% same-view targets reduces all four mean metrics relative to pure cross-view prediction. At $p_{\mathrm{same}}=.50$, fitted probing falls below both pure endpoints; structural transfer is lowest at $.75$ (Figure~\ref{fig:targetmix}; Table~\ref{tab:targetmix}). This non-monotonic response shows that the mixture need not inherit the strengths of either endpoint under the matched training budget. Final-checkpoint gradients characterize the resulting local geometry but do not explain the training dynamics (Appendix~\ref{app:objective}).

\begin{figure}[!htbp]
\centering
\includegraphics[width=\linewidth]{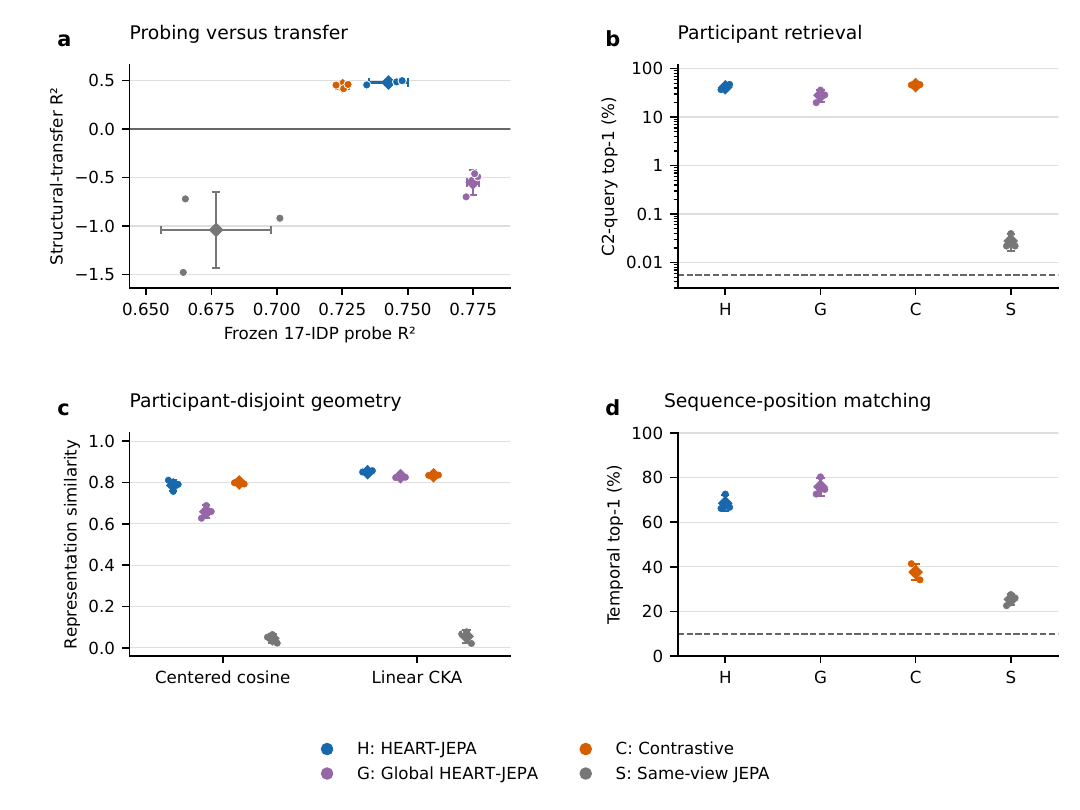}
\caption{\textbf{Different evaluation protocols capture distinct representation
properties.}
(a) Three-view fitted cardiac probing versus single-view structural transfer on the expanded analysis set;
bars show SD in both coordinates.
(b) Participant retrieval using C2 queries and the fixed 17{,}841-participant gallery.
(c) Centered cross-view representation geometry on C3.
(d) Sequence-position-centered temporal matching on C4.
Dots denote training seeds; diamonds and bars show mean $\pm$ sample SD.
Dashed lines indicate random baselines where applicable.
High geometric similarity or retrieval performance does not by itself imply
strong direct cross-view transfer.}
\label{fig:formal_diagnostics}
\end{figure}

\paragraph{Ancillary temporal matching.}
\label{sec:temporal}
Cross-view pretraining also improves sequence-position matching on C4 (Figure~\ref{fig:formal_diagnostics}d; Appendix~\ref{app:phase}). This measures acquisition indices, not synchronized physiological phases.

\subsection{Frozen features support ventricular function prediction}
\label{sec:ef}

Among seven models, HEART-JEPA has the lowest mean LVEF MAE
($\EFHeartLVEFMAEMean$ percentage points) and highest LVEF $R$
($\EFHeartLVEFRMean$; Table~\ref{tab:ef}). Relative to contrastive learning, LVEF MAE is lower by
\EFLVEFReductionVsContrastive\% and RVEF MAE by \EFRVEFReductionVsContrastive\%.
HEART-JEPA also improves both ventricular endpoints over same-view JEPA and the contextual pretrained comparators.
Global HEART-JEPA leads on RVEF (MAE $\EFGlobalRVEFMAEMean$ versus $\EFHeartRVEFMAEMean$), showing task-dependent strengths. External systems differ in pretraining; MR-JEPA's published EF results use a different cohort/input protocol and are outside this comparison \citep{jacob2026mrjepa}.

\begin{table}[!htbp]
\caption{Frozen-feature LVEF and RVEF prediction on the same 15{,}935 participants (C2), three-view inputs, and outer test folds. Values are mean $\pm$ sample SD over three five-fold CV repetitions. Metrics use pooled out-of-fold predictions, with checkpoint metrics averaged within each repetition. SD measures split sensitivity conditional on the evaluated checkpoints. MAE is in EF percentage points; $R$ is Pearson correlation. External models differ in pretraining. Bold marks the best mean; Appendix~\ref{app:ef_protocol} gives the aggregation.}
\label{tab:ef}
\centering
\footnotesize
\setlength{\tabcolsep}{2.0pt}
\renewcommand{\arraystretch}{1.18}
\resizebox{\linewidth}{!}{%
\begin{tabular}{@{}lccccc@{}}
\toprule
& & \multicolumn{2}{c}{LVEF} & \multicolumn{2}{c}{RVEF} \\
\cmidrule(lr){3-4}\cmidrule(lr){5-6}
Model & \shortstack{Pretrained\\checkpoints} &
MAE $\downarrow$ & $R\uparrow$ &
MAE $\downarrow$ & $R\uparrow$ \\
\midrule
\method{} & \EFHeartCheckpoints & $\mathbf{\EFHeartLVEFMAEMean} \pm \EFHeartLVEFMAESD$ & $\mathbf{\EFHeartLVEFRMean} \pm \EFHeartLVEFRSD$ & $\EFHeartRVEFMAEMean \pm \EFHeartRVEFMAESD$ & $\EFHeartRVEFRMean \pm \EFHeartRVEFRSD$ \\
Global HEART-JEPA & \EFGlobalCheckpoints & $\EFGlobalLVEFMAEMean \pm \EFGlobalLVEFMAESD$ & $\EFGlobalLVEFRMean \pm \EFGlobalLVEFRSD$ & $\mathbf{\EFGlobalRVEFMAEMean} \pm \EFGlobalRVEFMAESD$ & $\mathbf{\EFGlobalRVEFRMean} \pm \EFGlobalRVEFRSD$ \\
Same-view JEPA & \EFSameViewCheckpoints & $\EFSameViewLVEFMAEMean \pm \EFSameViewLVEFMAESD$ & $\EFSameViewLVEFRMean \pm \EFSameViewLVEFRSD$ & $\EFSameViewRVEFMAEMean \pm \EFSameViewRVEFMAESD$ & $\EFSameViewRVEFRMean \pm \EFSameViewRVEFRSD$ \\
\shortstack[l]{Cross-view\\contrastive} & \EFContrastiveCheckpoints & $\EFContrastiveLVEFMAEMean \pm \EFContrastiveLVEFMAESD$ & $\EFContrastiveLVEFRMean \pm \EFContrastiveLVEFRSD$ & $\EFContrastiveRVEFMAEMean \pm \EFContrastiveRVEFMAESD$ & $\EFContrastiveRVEFRMean \pm \EFContrastiveRVEFRSD$ \\
CMR Transformer & \EFCMRTransformerCheckpoints & $\EFCMRTransformerLVEFMAEMean \pm \EFCMRTransformerLVEFMAESD$ & $\EFCMRTransformerLVEFRMean \pm \EFCMRTransformerLVEFRSD$ & $\EFCMRTransformerRVEFMAEMean \pm \EFCMRTransformerRVEFMAESD$ & $\EFCMRTransformerRVEFRMean \pm \EFCMRTransformerRVEFRSD$ \\
CineMA & \EFCineMACheckpoints & $\EFCineMALVEFMAEMean \pm \EFCineMALVEFMAESD$ & $\EFCineMALVEFRMean \pm \EFCineMALVEFRSD$ & $\EFCineMARVEFMAEMean \pm \EFCineMARVEFMAESD$ & $\EFCineMARVEFRMean \pm \EFCineMARVEFRSD$ \\
ViTa-MAE & \EFViTaMAECheckpoints & $\EFViTaMAELVEFMAEMean \pm \EFViTaMAELVEFMAESD$ & $\EFViTaMAELVEFRMean \pm \EFViTaMAELVEFRSD$ & $\EFViTaMAERVEFMAEMean \pm \EFViTaMAERVEFMAESD$ & $\EFViTaMAERVEFRMean \pm \EFViTaMAERVEFRSD$ \\
\bottomrule

\end{tabular}%
}
\end{table}

\subsection{Downstream strengths vary by task}
\label{sec:contextual}

On C2, HEART-JEPA exceeds same-view JEPA in mean ROC-AUC and average precision (AP) for all five disease endpoints (Table~\ref{tab:disease_main}). Compared with contrastive learning, it has higher ROC-AUC for atrial fibrillation and heart failure and higher AP on four endpoints; the remaining comparisons favor contrastive learning. This internal comparison shares backbone, pretraining data, and splits. The disease model uses elementwise mean features, whereas continuous-IDP and EF probes concatenate views. Its results are specific to this aggregation; mean-versus-concatenation sensitivity is untested. Labels support discrimination, not prospective risk prediction (Appendix~\ref{app:ukb_disease}).

\begin{table}[!htbp]
\centering
\caption{\textbf{Conventional disease-label probing on C2.} Each cell is mean
ROC-AUC / AP. All five endpoints and all three evaluated internal objectives are
shown; means average folds and checkpoints within split, then three splits.
Appendix~\ref{app:ukb_disease} retains SDs, case counts, and the protocol.}
\label{tab:disease_main}
\footnotesize
\setlength{\tabcolsep}{6pt}
\begin{tabular}{@{}lccc@{}}
\toprule
Endpoint & HEART-JEPA & Same-view JEPA & Cross-view contrastive \\
\midrule
Atrial fibrillation & 0.777 / 0.291 & 0.752 / 0.267 & 0.776 / 0.301 \\
AV block & 0.797 / 0.131 & 0.753 / 0.071 & 0.807 / 0.103 \\
Diabetes & 0.773 / 0.176 & 0.758 / 0.151 & 0.776 / 0.172 \\
Heart failure & 0.801 / 0.162 & 0.752 / 0.121 & 0.786 / 0.136 \\
Myocardial infarction & 0.754 / 0.101 & 0.713 / 0.080 & 0.756 / 0.090 \\
\bottomrule
\end{tabular}
\end{table}

The separate benchmark checkpoint supports available-view and limited-label probing on the expanded set: three-view mean 17-IDP $R^2$ is $0.754\pm0.005$ across five folds, and two-view values range from 0.702 to 0.742. Using 1\%, 5\%, and 10\% of the labeled training pool gives scores of 0.587, 0.681, and 0.702, respectively, over five seeded 80/20 splits (Appendices~\ref{app:idp}--\ref{app:missing}).

On external M\&Ms-2, CMR Transformer has higher disease-versus-normal macro AUC (0.898 versus 0.848) and one-versus-rest macro AUC (0.801 versus 0.766). CineMA is also slightly higher (0.850 and 0.770). HEART-JEPA leads on hypertrophic cardiomyopathy versus normal and tetralogy of Fallot versus rest, rather than uniformly across tasks (Appendix~\ref{app:external}).

\section{Discussion and Limitations}

Cross-view target prediction supports reuse of a source-trained linear readout, a property not captured by fitted probing or retrieval alone. The target-refit control also reveals stronger accessible information in HEART-JEPA; the transfer advantage cannot be assigned solely to compatibility. When target features and labels are available, refitting is a reasonable alternative. Paired training views can reuse the same phenotype annotations, so these results do not establish annotation savings.

The matched sweep identifies sensitivity to target relation under the fixed training design. Objective-family controls change multiple factors, and final-checkpoint gradients do not identify the mechanism. Feature-moment normalization and orthogonal Procrustes test only two corrections; their failure does not exclude other alignments or prove view invariance. Seed SDs are descriptive, not confidence intervals.

Scope is limited to long-axis cine. Short-axis stacks alter input geometry; temporal matching is ancillary; and M\&Ms-2 supplies only 4Ch. External cross-view transfer and clinical missing-view utility remain untested. Conventional downstream benefits are task-dependent, reinforcing the value of reporting readout reuse alongside standard benchmarks.

\section{Conclusion}
\method{} improves fixed-readout transfer across spatially unaligned cardiac views compared with matched same-view pretraining. Target-refit and unlabeled correction controls bound this finding, identifying predictive compatibility as a useful complement to conventional downstream evaluation.

\clearpage
\section*{AI Use Statement}
We used generative AI tools for language editing and manuscript organization;
drafting cohort-selection and reproducibility descriptions from author-provided
study information; literature identification; clarification of the theoretical
argument; and feedback on methodology and result interpretation. They also
assisted with incorporating reviewer feedback, checking main-text and appendix
consistency, and preparing tables and figure labels from author-supplied summaries
and saved aggregate results. The authors are responsible for verifying this
content against study records, analysis outputs, and original sources and for
the accuracy and integrity of the manuscript.

\section*{Ethics Statement}
This study used de-identified UK Biobank and M\&Ms-2 data under their
respective access requirements. The models are for research use; clinical
validation is outside the present study.

UK Biobank data are available to approved researchers and cannot be redistributed
by the authors. M\&Ms-2 access follows the challenge conditions.

\section*{Reproducibility Statement}
Sections~\ref{sec:method}--\ref{sec:experiments} and Appendices~\ref{app:implementation}--\ref{app:audit} describe the
objective, architecture, preprocessing, and controlled training design. The paper
and appendices report aggregate results. Full replication additionally requires
the training and evaluation implementation and approved access to restricted UK
Biobank data. The predictive-compatibility protocol is described in Appendix~\ref{app:compatibility}.

\bibliography{references}
\bibliographystyle{iclr2027_conference}
\clearpage
\appendix
\raggedbottom
\setcounter{figure}{0}
\setcounter{table}{0}
\renewcommand{\thefigure}{S\arabic{figure}}
\renewcommand{\thetable}{S\arabic{table}}
\renewcommand{\theHfigure}{supp.\arabic{figure}}
\renewcommand{\theHtable}{supp.\arabic{table}}
\section{Why Cross-View Prediction Favors Context-Predictable Information}
\label{app:derivation}
The fixed-target squared-loss objective in Equation~\eqref{eq:token} has a
population-optimal predictor that averages out conditionally centered,
unpredictable target variation. We derive this context-predictability property under explicit
assumptions and distinguish it from direct transfer or alignment of encoder features.

\paragraph{The optimal fixed-target predictor.}
Fix an ordered view pair and a target position. Let $U$ contain the visible
context and the query information supplied to the predictor, and let $Y$ be
the corresponding fixed, square-integrable target feature. With
$m(U)=\mathbb{E}[Y\mid U]$, any square-integrable predictor $f$ satisfies
\begin{equation}
\mathbb{E}\|Y-f(U)\|_2^2
=\mathbb{E}\|Y-m(U)\|_2^2
 +\mathbb{E}\|m(U)-f(U)\|_2^2.
\label{eq:risk_decomposition}
\end{equation}
To see this, expand $Y-f(U)=(Y-m(U))+(m(U)-f(U))$.
The cross term is zero because $\mathbb{E}[Y-m(U)\mid U]=0$.
Hence $f^*(U)=\mathbb{E}[Y\mid U]$ minimizes the population loss.

Suppose the fixed target can be written as
\begin{equation}
Y=A_t C+B_t S_t+\epsilon_t,
\label{eq:fixed_target_decomposition}
\end{equation}
where $C$ contains information shared across views, $S_t$ is target-specific
variation, and $A_t,B_t$ are fixed linear maps. Assume
$S_t\perp U\mid C$, $\mathbb{E}[S_t\mid C]=0$, and
$\mathbb{E}[\epsilon_t\mid U]=0$. The tower property gives
\begin{equation}
\mathbb{E}[S_t\mid U]
=\mathbb{E}\!\left[\mathbb{E}[S_t\mid U,C]\mid U\right]
=\mathbb{E}\!\left[\mathbb{E}[S_t\mid C]\mid U\right]=0.
\label{eq:residual_centering}
\end{equation}
It follows that
\begin{equation}
f^*(U)=A_t\mathbb{E}[C\mid U].
\label{eq:optimal_crossview_predictor}
\end{equation}
Under these assumptions, the optimal predictor retains the context-predictable
part of the shared state and averages out unpredictable target-specific
variation. The assumptions apply to each view pair and queried position,
including the masking and query-selection rule. Acquisition factors that are
predictable across views can be retained.

\paragraph{Encoder compatibility is a separate property.}
Let two views represent the same scalar by $z^c=s$ and $z^t=-s$. A predictor
can translate perfectly between them, but a source-view prediction model
$f_c(z)=z$ fails when applied unchanged to $z^t$. Thus, accurate cross-view
latent prediction does not guarantee accurate predictions under the transfer
protocol in Equation~\eqref{eq:cross_view_transfer}.
The derivation holds the target fixed, whereas EMA training updates it jointly
with the online encoder. Constant representations and information unused by
the predictor remain compatible with the fixed-target analysis. We therefore
evaluate encoder compatibility empirically through cardiac prediction,
retrieval, geometry, and view classification.

\section{Implementation Details}
\label{app:implementation}

\paragraph{Architecture and inputs.}
The encoder uses 12 Transformer blocks, six attention heads, and 384-dimensional tokens.
The predictor uses six Transformer blocks, dimension 128, and four heads.
The predictor takes 384-dimensional context tokens and uses a 128-dimensional
internal representation. Its output in Equation~\eqref{eq:token} is required
to match the 384-dimensional target tokens.
Each input has
50 frames at $128\times128$ resolution, with spatial patch size 16 and temporal tube size
5, yielding a native $10\times8\times8$ token grid. MultiBlockMask3D samples eight target
blocks with spatial scale $[0.15,0.45]$, aspect ratio $[0.75,1.5]$, and temporal scale
$[0.5,1.0]$. The EMA coefficient is 0.996. The target-view identity is supplied only to
the predictor. The full model contains approximately 44.9M parameters, of which 23.1M are
optimized directly; downstream evaluations freeze the EMA target encoder. Formal runs use
791 optimizer updates per epoch and 276{,}850 updates in total, with a constant learning
rate of $1.5\times10^{-4}$ and no warmup or decay. The same schedule and
optimization budget apply to every condition. Video preprocessing retains
the first 50 frames or repeats the last frame when fewer are available; it does not
temporally interpolate or register cardiac phases. Crops use the union of nonzero
segmentation pixels across frames, and each cropped frame is z-scored before bilinear
resizing. The ViTa-MAE comparator is its image-only
checkpoint; the evaluation adapter repeats a long-axis view across its nine expected
spatial slices. This adapter is part of the external-model evaluation protocol.
All controlled objectives share the segmentation-derived crop prior.
Padding prevalence and sensitivity to temporal resampling remain unmeasured.

\paragraph{Context masks and target queries.}
Let $M_c$ be the union of eight sampled MultiBlockMask3D regions. Spatial and temporal positional embeddings are added to all tokens before the context sequence is restricted to $\mathcal G\setminus M_c$. In the primary model, the predictor queries and the target features selected for the loss use $Q_t = M_c$. These shared indices identify positions on separately normalized $10\times8\times8$ lattices; the target encoder processes the full video without explicit plane registration. Each update samples one ordered view pair and one $M_c$ for the minibatch. The independent-query control obtains $Q_t$ via a second independent draw from the same masking distribution. Across 10{,}000 fixed-seed audit draws, the two sets were never identical: they contained $470.97 \pm 47.62$ and $470.61 \pm 47.36$ tokens, respectively, with a mean Jaccard overlap of $0.682 \pm 0.072$. This intervention removes exact set coupling while preserving incidental overlap and the marginal target-count distribution. The cardinalities $|Q_t|$ and $|M_c|$ can differ within an update.

\paragraph{Global HEART-JEPA.}
Global HEART-JEPA is the global non-contrastive variant of \method{}. It applies
cross-view latent prediction to one pooled video representation instead of a
set of target tokens. The online branch mean-pools the visible context tokens
from the masked context view and adds a learned embedding specifying the target
view. The resulting representation is passed through an online projector and
prediction MLP. In the target branch, all tokens from the full target view are
mean-pooled and passed through an EMA-updated copy of the projector. The loss is
the squared Euclidean distance between the $\ell_2$-normalized 128-dimensional
online prediction and the stop-gradient target projection, without negative
samples. EMA updates the target projector parameters; its BatchNorm layers use target-batch
statistics and maintain branch-specific running buffers. The stop-gradient
operation excludes the target branch from backpropagation.

To control for head capacity, we set the projector and predictor hidden
dimensions to 1,649 and 1,743, respectively, yielding 1,298,688 trainable
parameters in the online head, exactly matching the token predictor. The
additional EMA projector increases the total parameter count from 44.9M to
45.8M. Predicting one representation per video reduces head-level FLOPs
relative to predicting a set of target tokens.

Using the encoder notation in Section~\ref{sec:architecture}, let
$h^c=E_\theta(x^{v_c};\mathcal G\setminus M_c)$ and
$H^t=E_\phi(x^{v_t};\mathcal G)$. Write $\operatorname{pool}(\cdot)$ for
averaging the supplied tokens and $\operatorname{norm}(u)=u/\lVert u\rVert_2$.
Writing both outputs in the same 128-dimensional projection space, the loss is
\begin{equation}
\mathcal L_{\mathrm{global}}
=\left\|\operatorname{norm}\!\left(
q_\psi\!\left(g_\psi\!\left(\operatorname{pool}(h^c)+e_{v_t}\right)\right)
\right)
-\sg\!\left[\operatorname{norm}\!\left(
g_\xi\!\left(\operatorname{pool}(H^t)\right)
\right)\right]\right\|_2^2.
\label{eq:global}
\end{equation}
Here $g_\psi$ and $q_\psi$ are the online projector and predictor,
$g_\xi$ is the EMA target projector, and $e_{v_t}$ is the target-view embedding.
Context pooling uses only visible tokens; target pooling uses the full target grid.

\paragraph{Cross-view contrastive learning.}
The EMA-InfoNCE control instead discriminates participants: the masked context
is the query, a full distinct view of the same participant is its positive
key, and other participants in the minibatch provide negative keys. It shares
the backbone and training protocol but uses a global projection head rather
than the token predictor. These objective-family comparisons jointly vary pooling, normalization,
head design, and the use of negatives.

\section{Training Design and Cohort Separation}
\label{app:audit}

The target-mixture experiment changes only $p_{\mathrm{same}}$ within each
training seed. All five settings share the manifest, ordered study split,
architecture, masking, preprocessing, optimizer, update count, and checkpoint
rule. Seeds 42, 43, and 44 give three stochastic training runs on the same
ordered split (split seed 42). Table~\ref{tab:full_config_audit} summarizes
what is held constant and what changes in the additional controls.

\begin{table}[H]
\caption{Controlled training design. The formal mixture includes
$p_{\mathrm{same}}\in\{0,.25,.5,.75,1\}$. All columns use the same manifest,
ordered split, deterministic preprocessing, batch size 48, AdamW, learning rate
$1.5\times10^{-4}$, weight decay 0.05, constant schedule, EMA coefficient 0.996,
276,850 updates.}
\label{tab:full_config_audit}
\centering
\footnotesize
\setlength{\tabcolsep}{3pt}
\begin{tabular}{@{}lcccc@{}}
\toprule
 & Target mixture & Independent queries & Global HEART-JEPA & Contrastive \\
\midrule
Encoder & ViT-12 & ViT-12 & ViT-12 & ViT-12 \\
Prediction unit & Target tokens & Target tokens & Pooled video & Pooled video \\
Objective & Token MSE & Token MSE & Normalized regression & EMA-InfoNCE \\
Trainable parameters & 23.1M & 23.1M & 23.1M & 22.1M \\
Total parameters & 44.9M & 44.9M & 45.8M & --- \\
Context mask & MultiBlockMask3D & Same distribution & Same distribution & Same distribution \\
Target query indices & $M_c$ & Independent $Q_t$ & Not applicable & Not applicable \\
Training seeds & 42, 43, 44 & 42, 43, 44 & 42, 43, 44 & 42, 43, 44 \\
\bottomrule
\end{tabular}
\end{table}
The independent-query control changes only the target-query draw. Global HEART-JEPA
matches the number of trainable head parameters, but its EMA projector changes
the total parameter count and its single prediction reduces head-level compute.
The contrastive head has fewer trainable parameters. These controls compare
objective designs with different head-level compute as well as prediction granularity.

\paragraph{Checkpoint scope across evaluations.}
The EF benchmark is reported separately from the matched intervention.
Table~\ref{tab:ef} lists three checkpoints each for \method{}, Global HEART-JEPA,
same-view JEPA, and cross-view contrastive learning; each external model uses
one. The retained provenance does not establish a one-to-one mapping between these EF checkpoints and the matched intervention runs with seeds 42--44, so Table~\ref{tab:ef} is not interpreted as a seed-matched extension of the intervention. The available-view, label-efficiency, and external analyses use a separate single benchmark checkpoint, distinct from the matched three-run means.

\paragraph{Evaluation cohort and expanded analysis set.}
C2 contains 15{,}935 participants absent from both pretraining subsets and is
the primary participant-disjoint evaluation cohort. The expanded analysis set
contains 17{,}841 participants with no exact study overlap at the visit and
instance level; 1{,}906 have another visit in pretraining. Excluding these
participants yields C2. The full target-mixture sweep, objective-control
probing and transfer, and contextual frozen-feature benchmarks use the expanded
set. The fixed retrieval gallery also contains 17{,}841 participants; each
analysis specifies whether queries use C2 or the full expanded set. The
study-level training/monitoring split includes 127 participants with different
visits on both sides.

\begin{table}[H]
\caption{Matched target-mixture intervention. Values are mean $ \pm $ sample SD over three
training seeds, on the expanded analysis set (17{,}841 participants). Retrieval is top-1
percentage over six ordered view directions, with queries and gallery both drawn from
that set; transfer is cross-view structural \Rtwo{} without target-view refitting;
cardiac probing uses all 17 IDPs.}
\label{tab:targetmix}
\centering
\footnotesize
\begin{tabular}{ccccc}
\toprule
$p_{\mathrm{same}}$ & Matched cosine & Transfer \Rtwo{} & Retrieval top-1 & Cardiac probe \Rtwo{} \\
\midrule
0 & $\FormalPZeroCosineMean \pm \FormalPZeroCosineSD$ & $\FormalPZeroStructuralTransferMean \pm \FormalPZeroStructuralTransferSD$ & $\FormalPZeroRetrievalMean \pm \FormalPZeroRetrievalSD$ & $\FormalPZeroMorphologyMean \pm \FormalPZeroMorphologySD$ \\
.25 & $\FormalPTwentyFiveCosineMean \pm \FormalPTwentyFiveCosineSD$ & $\FormalPTwentyFiveStructuralTransferMean \pm \FormalPTwentyFiveStructuralTransferSD$ & $\FormalPTwentyFiveRetrievalMean \pm \FormalPTwentyFiveRetrievalSD$ & $\FormalPTwentyFiveMorphologyMean \pm \FormalPTwentyFiveMorphologySD$ \\
.50 & $\FormalPFiftyCosineMean \pm \FormalPFiftyCosineSD$ & $\FormalPFiftyStructuralTransferMean \pm \FormalPFiftyStructuralTransferSD$ & $\FormalPFiftyRetrievalMean \pm \FormalPFiftyRetrievalSD$ & $\FormalPFiftyMorphologyMean \pm \FormalPFiftyMorphologySD$ \\
.75 & $\FormalPSeventyFiveCosineMean \pm \FormalPSeventyFiveCosineSD$ & $\FormalPSeventyFiveStructuralTransferMean \pm \FormalPSeventyFiveStructuralTransferSD$ & $\FormalPSeventyFiveRetrievalMean \pm \FormalPSeventyFiveRetrievalSD$ & $\FormalPSeventyFiveMorphologyMean \pm \FormalPSeventyFiveMorphologySD$ \\
1 & $\FormalPOneCosineMean \pm \FormalPOneCosineSD$ & $\FormalPOneStructuralTransferMean \pm \FormalPOneStructuralTransferSD$ & $\FormalPOneRetrievalMean \pm \FormalPOneRetrievalSD$ & $\FormalPOneMorphologyMean \pm \FormalPOneMorphologySD$ \\
\bottomrule

\end{tabular}
\end{table}

\begin{table}[H]
\caption{Per-seed results for the endpoints of the matched target-source intervention.
Retrieval is top-1 percentage; transfer and cardiac probing are \Rtwo{}.}
\label{tab:targetmix_seeds}
\centering
\footnotesize
\begin{tabular}{ccrrrr}
\toprule
$p_{\mathrm{same}}$ & Seed & Matched cosine & Transfer & Retrieval & Cardiac probe \\
\midrule
0 & 42 & 0.918 & 0.454 & 36.7683 & 0.734 \\
0 & 43 & 0.956 & 0.485 & 39.3205 & 0.746 \\
0 & 44 & 0.923 & 0.497 & 47.2937 & 0.748 \\
1 & 42 & 0.081 & $-1.478$ & 0.0234 & 0.664 \\
1 & 43 & 0.215 & $-0.919$ & 0.0392 & 0.701 \\
1 & 44 & 0.124 & $-0.720$ & 0.0206 & 0.665 \\
\bottomrule

\end{tabular}
\end{table}

\section{Target-Query and Global HEART-JEPA Results}
\label{app:target_relation_controls}

Independent target queries preserve positive structural transfer and similar
cardiac probing, with higher functional transfer in all three paired seeds
(Tables~\ref{tab:target_relation_controls} and
\ref{tab:target_relation_control_seeds}). All results use frozen EMA encoders;
probing and transfer use the expanded analysis set, and retrieval uses C2
queries against the fixed 17{,}841-participant gallery. The intervention removes exact context-mask index reuse while
retaining incidental overlap (Jaccard $0.682\pm0.072$).

Global HEART-JEPA has stronger fitted cardiac probing and higher raw cosine
similarity in all three seeds, with a matched--mismatched margin of
$\TargetControlGlobalMarginMean\pm\TargetControlGlobalMarginSD$.
\method{} has a larger margin and better participant retrieval, structural transfer, and functional
transfer. Useful cardiac information therefore coexists with negative mean
transfer in Global HEART-JEPA. Its stronger RVEF prediction in
Table~\ref{tab:ef} further distinguishes fitting a new prediction model from
transferring one unchanged. Head design and normalization vary jointly in
this comparison, leaving their individual contributions unresolved.

\begin{figure}[!htbp]
\centering
\includegraphics[width=\linewidth]{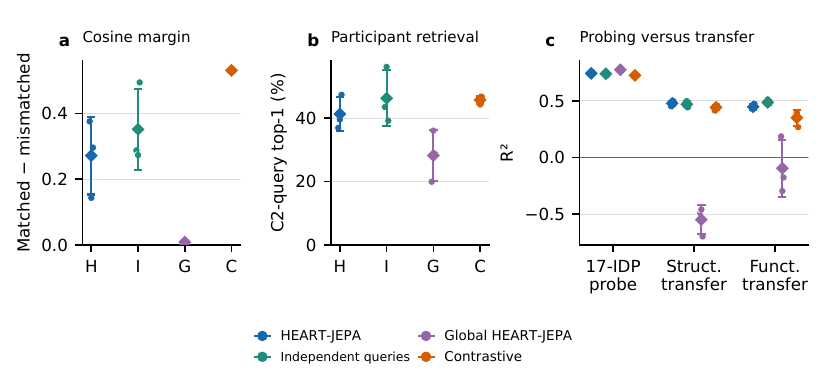}
\caption{Cross-view objective controls. H: shared-query \method{}; I: independent target
queries; G: Global HEART-JEPA; C: EMA-InfoNCE. Points are three training
seeds and diamonds are mean $ \pm $ sample SD. Margin and cardiac probes use the expanded
analysis set; retrieval uses C2 queries with the fixed 17{,}841-participant gallery. The Global HEART-JEPA control has the
higher 17-IDP probe score but negative mean structural and functional transfer. Head design
and normalization differ between objective families.}
\label{fig:objective_controls}
\end{figure}

\begin{table}[H]
\caption{Target-relation controls (mean $ \pm $ sample SD over seeds 42--44). Retrieval is
top-1 percentage; cardiac probing and transfer are \Rtwo{}.}
\label{tab:target_relation_controls}
\centering
\footnotesize
\setlength{\tabcolsep}{2.6pt}
\begin{tabular}{lrrrrr}
\toprule
Objective & \shortstack{Matched\\cosine} & \shortstack{Retrieval\\(\%)} & \shortstack{Cardiac\\probe} & \shortstack{Structural\\transfer} & \shortstack{Functional\\transfer} \\
\midrule
HEART-JEPA & $\TargetControlHeartCosineMean \pm \TargetControlHeartCosineSD$ & $\TargetControlHeartRetrievalMean \pm \TargetControlHeartRetrievalSD$ & $\TargetControlHeartMorphologyMean \pm \TargetControlHeartMorphologySD$ & $\TargetControlHeartStructuralTransferMean \pm \TargetControlHeartStructuralTransferSD$ & $\TargetControlHeartFunctionalTransferMean \pm \TargetControlHeartFunctionalTransferSD$ \\
Independent queries & $\TargetControlIndependentCosineMean \pm \TargetControlIndependentCosineSD$ & $\TargetControlIndependentRetrievalMean \pm \TargetControlIndependentRetrievalSD$ & $\TargetControlIndependentMorphologyMean \pm \TargetControlIndependentMorphologySD$ & $\TargetControlIndependentStructuralTransferMean \pm \TargetControlIndependentStructuralTransferSD$ & $\TargetControlIndependentFunctionalTransferMean \pm \TargetControlIndependentFunctionalTransferSD$ \\
Global HEART-JEPA & $\TargetControlGlobalCosineMean \pm \TargetControlGlobalCosineSD$ & $\TargetControlGlobalRetrievalMean \pm \TargetControlGlobalRetrievalSD$ & $\TargetControlGlobalMorphologyMean \pm \TargetControlGlobalMorphologySD$ & $\TargetControlGlobalStructuralTransferMean \pm \TargetControlGlobalStructuralTransferSD$ & $\TargetControlGlobalFunctionalTransferMean \pm \TargetControlGlobalFunctionalTransferSD$ \\
\bottomrule

\end{tabular}
\end{table}

\begin{table}[H]
\caption{Per-seed target-relation control results. Ret. and View ID are percentages;
Probe, Struct., and Funct. are \Rtwo{}. H denotes \method{} with shared query indices, I denotes
independent target queries, and G denotes Global HEART-JEPA.}
\label{tab:target_relation_control_seeds}
\centering
\footnotesize
\setlength{\tabcolsep}{3pt}
\begin{tabular*}{\linewidth}{@{\extracolsep{\fill}}lcrrrrrrr@{}}
\toprule
Control & Seed & Cosine & Margin & Ret. & View ID & Probe & Struct. & Funct. \\
\midrule
H & 42 & 0.918 & 0.3761 & 36.87 & 99.976 & 0.734 & 0.454 & 0.430 \\
H & 43 & 0.956 & 0.1430 & 39.52 & 99.979 & 0.746 & 0.485 & 0.438 \\
H & 44 & 0.923 & 0.2961 & 47.32 & 99.978 & 0.748 & 0.497 & 0.474 \\
I & 42 & 0.932 & 0.2878 & 43.47 & 99.974 & 0.742 & 0.469 & 0.475 \\
I & 43 & 0.930 & 0.2735 & 56.10 & 99.981 & 0.748 & 0.498 & 0.509 \\
I & 44 & 0.907 & 0.4949 & 39.15 & 99.979 & 0.728 & 0.445 & 0.475 \\
G & 42 & 0.987 & 0.0067 & 19.88 & 99.989 & 0.777 & $-0.492$ & 0.186 \\
G & 43 & 0.991 & 0.0073 & 36.00 & 99.989 & 0.776 & $-0.460$ & $-0.297$ \\
G & 44 & 0.985 & 0.0099 & 28.65 & 99.991 & 0.772 & $-0.699$ & $-0.177$ \\
\bottomrule

\end{tabular*}
\end{table}

\section{Matched Cross-View Contrastive Control}
\label{app:contrastive}

The three contrastive runs are paired by training seed with the pure
cross-view \method{} runs. Retrieval uses C2 queries and the fixed
17{,}841-participant gallery; cardiac probing and direct transfer use the
expanded analysis set. The comparison
asks whether an objective optimized for participant discrimination also
supports cardiac prediction across views.

\begin{table}[H]
\caption{Matched objective-family comparison (mean $ \pm $ sample SD over seeds 42--44).
Retrieval is top-1 percentage; cardiac probing and transfer are \Rtwo{}.}
\label{tab:contrastive_summary}
\centering
\footnotesize
\setlength{\tabcolsep}{2.6pt}
\begin{tabular}{lrrrrr}
\toprule
Objective & Margin & Retrieval & Cardiac probe & Structural & Functional \\
\midrule
\method{}
& $\HeartObjectiveMarginMean \pm \HeartObjectiveMarginSD$
& $\HeartObjectiveRetrievalMean \pm \HeartObjectiveRetrievalSD$
& $\HeartObjectiveMorphologyMean \pm \HeartObjectiveMorphologySD$
& $\HeartObjectiveStructuralTransferMean \pm \HeartObjectiveStructuralTransferSD$
& $\HeartObjectiveFunctionalTransferMean \pm \HeartObjectiveFunctionalTransferSD$ \\
Contrastive
& $\ContrastiveMarginMean \pm \ContrastiveMarginSD$
& $\ContrastiveRetrievalMean \pm \ContrastiveRetrievalSD$
& $\ContrastiveMorphologyMean \pm \ContrastiveMorphologySD$
& $\ContrastiveStructuralTransferMean \pm \ContrastiveStructuralTransferSD$
& $\ContrastiveFunctionalTransferMean \pm \ContrastiveFunctionalTransferSD$ \\
\bottomrule
\end{tabular}
\end{table}

\method{} has higher matched cosine in all three paired seeds
($\HeartObjectiveCosineMean \pm \HeartObjectiveCosineSD$ versus
$\ContrastiveCosineMean \pm \ContrastiveCosineSD$), while contrastive learning has the larger
margin in all three. \method{} has higher cardiac-probe, structural-transfer, and
functional-transfer scores in all three; contrastive retrieval is higher in two. View identity
is nearly identical at
$\HeartObjectiveViewIdentityMean \pm \HeartObjectiveViewIdentitySD\%$ and
$\ContrastiveViewIdentityMean \pm \ContrastiveViewIdentitySD\%$.

\section{Participant-Disjoint Transfer and Representation Geometry}
\label{app:strict}

Table~\ref{tab:strict_endpoints} reports the formal endpoints after excluding every
evaluation participant present in the pretraining manifest. Centered geometry uses C3, identical
across models; cardiac probes use C2, and retrieval uses C2 queries with the fixed 17{,}841-participant gallery. Values are means and sample SD across seeds
42--44. Centering removes view-specific means for the geometric diagnostic;
the transfer pipeline retains source-fitted standardization. Linear CKA
characterizes full-feature geometry independently of a fitted target-view mapping.
A linear source probe has predictions $w_c^\top D_c^{-1}(z^t-\mu_c)+b_c$,
where $D_c$ is its diagonal source scale matrix. A target feature-mean shift
adds $w_c^\top D_c^{-1}(\mu_t-\mu_c)$ to the prediction mean and can reduce
\Rtwo{} without changing Pearson correlation under a constant output shift.
Transferred $R^2$ therefore combines feature-moment calibration and predictive-direction effects. This is intentional for the unchanged-model protocol: source-view standardization is part of the mapping whose cross-view validity is being tested. We do not interpret negative transferred $R^2$ as absence of phenotype information. The additional C2 diagnostics in Appendix~\ref{app:compatibility} now measure transferred Pearson correlation, per-feature target moments, paired Procrustes, and target-specific refitting. They test specific explanations without identifying a unique mechanism.

On C2, linear view classification is
$\StrictPZeroViewIdentityMean\pm\StrictPZeroViewIdentitySD\%$ for cross-view
pretraining and $\StrictPOneViewIdentityMean\pm\StrictPOneViewIdentitySD\%$
for same-view pretraining, preserving view identity in both representations.

\begin{table}[H]
\caption{Participant-disjoint (C2) endpoint comparison. Retrieval is top-1 percentage with C2 queries
and the fixed 17{,}841-participant gallery; transfer and cardiac probing are \Rtwo{}. Linear CKA is invariant
to orthogonal transforms and isotropic scaling.}
\label{tab:strict_endpoints}
\centering
\footnotesize
\setlength{\tabcolsep}{3.0pt}
\begin{tabular}{lrrrrr}
\toprule
Target & \shortstack{Centered cosine\\C3} & \shortstack{Linear CKA\\C3} & \shortstack{Retrieval\\C2 queries} & \shortstack{Structural transfer\\C2} & \shortstack{Cardiac probe\\C2} \\
\midrule
Cross-view & $\StrictPZeroCenteredCosineMean \pm \StrictPZeroCenteredCosineSD$
& $\StrictPZeroCKAMean \pm \StrictPZeroCKASD$
& $\StrictPZeroRetrievalMean \pm \StrictPZeroRetrievalSD$
& $\StrictPZeroStructuralTransferMean \pm \StrictPZeroStructuralTransferSD$
& $\StrictPZeroMorphologyMean \pm \StrictPZeroMorphologySD$ \\
Same-view & $\StrictPOneCenteredCosineMean \pm \StrictPOneCenteredCosineSD$
& $\StrictPOneCKAMean \pm \StrictPOneCKASD$
& $\StrictPOneRetrievalMean \pm \StrictPOneRetrievalSD$
& $\StrictPOneStructuralTransferMean \pm \StrictPOneStructuralTransferSD$
& $\StrictPOneMorphologyMean \pm \StrictPOneMorphologySD$ \\
\bottomrule
\end{tabular}
\end{table}

\section{Predictive-Compatibility Diagnostics}
\label{app:compatibility}

\paragraph{Fixed cohort and reproduction.}
The diagnostics retain C2 (15{,}935 participants excluded from the pretraining
manifest), its ordered five participant folds, the phenotype schema, and frozen
features. The primary diagnostic panel is the original 13 structural traits. The evaluation
grid also includes the four functional traits; the existing functional-transfer
results are retained separately in the original endpoint tables. HEART-JEPA and
same-view JEPA form the matched target-relation comparison; cross-view contrastive
is a contextual objective control. Across six directions, five folds, 17 traits,
and three pretrained seeds, the matched direct grid has 3,060 rows. These match the cached evaluation exactly;
refitting the original source pipeline reproduces every direct-transfer $R^2$
with maximum absolute error $1.33\times10^{-15}$. This reproduction used one BLAS
thread because float32 Cholesky calculations were thread-order sensitive.

\paragraph{Four readout conditions and direct Pearson correlation.}
On each fold, the source probe is a StandardScaler followed by multi-output
Ridge($\alpha=1$, solver=Cholesky), fitted on source training participants with all
17 labels complete. The same fitted source ridge, including its intercept, is
used in the first three conditions below. All fitting uses training participants
only; no held-out participant features or labels enter a scaler, map, or ridge.
With row-wise features, write $u^v=S_v(z^v)$:
\begin{enumerate}
\item \textbf{Direct:} $f_c(S_c(z^t))$, with the source scaler and ridge unchanged.
Pearson $r$ is calculated from these same direct predictions and the held-out
phenotypes; it is not a separately calibrated prediction or a rank correlation.
\item \textbf{Target moments:} $f_c(S_t(z^t))$, where $S_t$ uses only target
training-feature means and per-feature scales, without labels. This is feature
normalization, not output calibration or general distribution alignment.
\item \textbf{Paired Procrustes:} $f_c(S_t(z^t)Q^*)$, where
$Q^*=\arg\min_{Q^\top Q=I}\|U_tQ-U_c\|_F^2$ is fitted to paired standardized
training features, without phenotype labels. Training-pair Frobenius discrepancy
decreased in all 270 model $\times$ direction $\times$ fold $\times$ seed mappings.
This tests one orthogonal correction, not arbitrary learned alignment.
\item \textbf{Target refit:} $f_t(S_t(z^t))$, with both scaler and ridge fitted
on target training features and labels. This measures linearly accessible
information under the chosen probe and sample size, rather than total information,
a label-free correction, or a strict upper bound for arbitrary supervised models.
\end{enumerate}
All conditions retain a supervised phenotype readout. The two unlabeled corrections
do not use labels to fit their scalers or maps. Because training observations are
paired, target refitting can reuse the source participants' phenotype annotations;
the distinction is a new target-feature fit, not necessarily new annotation.

\begin{figure}[!htbp]
\centering
\includegraphics[width=\linewidth]{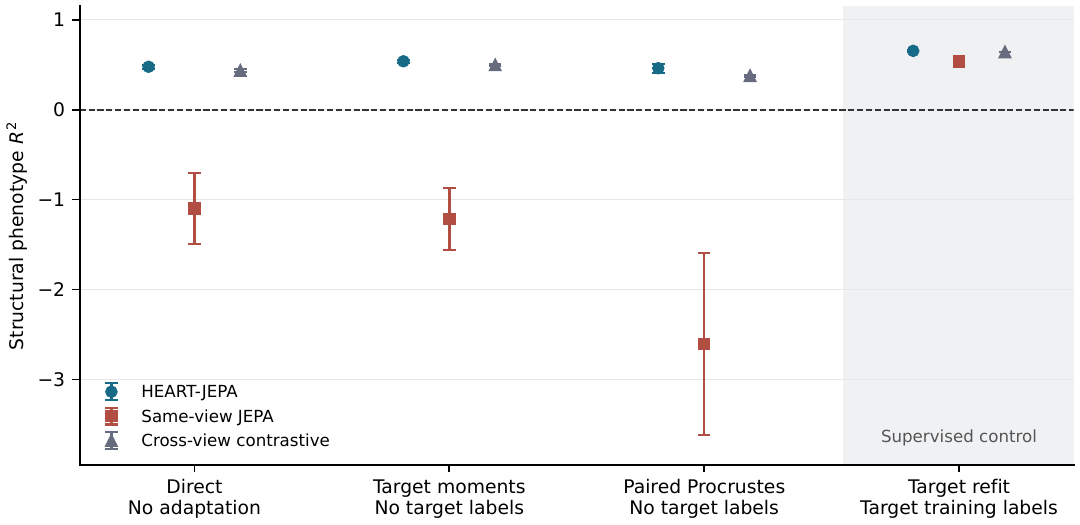}
\caption{\textbf{Readout conditions for structural transfer on C2.}
Points are means and bars are sample SD across three pretrained seeds, averaging
13 traits, six directions, and five folds within seed. The complete negative range
and zero reference are shown. Conditions use increasing types of target-view
information but are distinct evaluations, not a guaranteed improvement path;
target refit uses labels and is marked separately. Points and bars use the reported three-decimal aggregate means and SDs; individual
seed observations are not displayed.}
\label{fig:compatibility_conditions}
\end{figure}

\paragraph{Aggregation and interpretation.}
Each primary seed mean averages 13 structural traits $\times$ six ordered
directions $\times$ five folds. Table~\ref{tab:compatibility} and
Figure~\ref{fig:compatibility_conditions} show mean $\pm$ sample SD over three
pretrained seeds. The conditions differ in target-data and label access and need
not improve monotonically. Folds, traits, and directions are correlated; no
significance test is inferred from these seed summaries.
The within-seed target-refit minus direct $R^2$ gap is $0.177\pm0.010$ for
HEART-JEPA, $1.638\pm0.381$ for same-view JEPA, and $0.205\pm0.019$ for
contrastive learning. Moment normalization changes direct $R^2$ by
$+0.061\pm0.005$, $-0.115\pm0.108$, and $+0.062\pm0.024$, respectively.
These paired summaries use unrounded seed values, not subtraction of rounded
table entries or unpaired uncertainty intervals. The target-specific difference
prevents attributing the full transfer gain to compatibility alone.

\begin{table}[H]
\centering
\caption{Mean direct structural $R^2$ by ordered direction on C2, averaged across
traits, folds, and seeds. Values are descriptive direction means; across-seed SDs for the overall summary are in Table~\ref{tab:compatibility}.}
\label{tab:compatibility_directions}
\small
\begin{tabular}{lrrr}
\toprule
Direction & HEART-JEPA & Same-view JEPA & Cross-view contrastive \\
\midrule
2Ch $\rightarrow$ 3Ch & 0.423 & $-2.630$ & 0.432 \\
2Ch $\rightarrow$ 4Ch & 0.493 & $-1.634$ & 0.449 \\
3Ch $\rightarrow$ 2Ch & 0.500 & $-0.725$ & 0.415 \\
3Ch $\rightarrow$ 4Ch & 0.547 & $-0.271$ & 0.515 \\
4Ch $\rightarrow$ 2Ch & 0.483 & $-0.901$ & 0.402 \\
4Ch $\rightarrow$ 3Ch & 0.417 & $-0.447$ & 0.396 \\
\bottomrule
\end{tabular}
\end{table}

HEART-JEPA has positive direction means in all six directions, while same-view
JEPA has negative means in all six. HEART-JEPA has higher
direct-transfer means for all 13 structural traits versus same-view JEPA and
12 of 13 versus contrastive learning. These counts are descriptive, not
independent tests. RA maximum/minimum volume remain weak absolute endpoints
($R^2=0.112/0.162$). Thus, the result is not uniformly strong across anatomy.

\paragraph{Source-data availability.}
The paper reports structural diagnostic aggregates. Full replication requires the
training and evaluation implementation and approved access to restricted data.

\section{Temporal Matching Across Sequence Positions}
\label{app:phase}

Ten temporal tubelets partition each fixed-length 50-frame input into normalized sequence
positions. Inputs are truncated or repeat-last-padded, not temporally resampled. Sequence-position centering subtracts the across-participant mean token separately for each view and
position. Table~\ref{tab:phase_endpoints} averages each metric over six ordered view pairs
within a seed and then reports mean $ \pm $ sample SD over three seeds. The circular MAE is
normalized by ten positions and treats sequence indices cyclically. It measures
sequence-index correspondence rather than physiological phase error. Repeated
terminal frames are a potential matching cue; padding prevalence and a
phase-resampled control remain unmeasured. We therefore treat temporal matching as an ancillary representation diagnostic rather than evidence of physiological phase alignment or a component of the main direct-transfer claim. The relationship between this matching result and EF prediction has not been tested.

\begin{table}[H]
\caption{Sequence-position-centered retrieval on C4. Top-1 chance is 10\%.}
\label{tab:phase_endpoints}
\centering
\footnotesize
\begin{tabular}{lrrrr}
\toprule
Target & Top-1 (\%) & Circular MAE & Same-index cosine & Index margin \\
\midrule
Cross-view & $\PhasePZeroTopOneMean \pm \PhasePZeroTopOneSD$
& $\PhasePZeroCircularMAEMean \pm \PhasePZeroCircularMAESD$
& $\PhasePZeroCosineMean \pm \PhasePZeroCosineSD$
& $\PhasePZeroMarginMean \pm \PhasePZeroMarginSD$ \\
Same-view & $\PhasePOneTopOneMean \pm \PhasePOneTopOneSD$
& $\PhasePOneCircularMAEMean \pm \PhasePOneCircularMAESD$
& $\PhasePOneCosineMean \pm \PhasePOneCosineSD$
& $\PhasePOneMarginMean \pm \PhasePOneMarginSD$ \\
\bottomrule
\end{tabular}
\end{table}

\section{Final-Checkpoint Objective Diagnostics}
\label{app:objective}

To characterize the sharp change near $p_{\mathrm{same}}=.25$ discussed in
Section~\ref{sec:mixture}, we evaluate each final checkpoint on fixed held-out batches
under all same-view and cross-view relations, normalizing MSE by target-token variance. On
a smaller fixed batch we also compute the cosine similarity between gradients induced by
the same-view and cross-view losses. The pure cross-view endpoint has lower held-out
cross-view error and more aligned same/cross local gradients. The measurements
characterize local loss geometry at the final solution. Near-orthogonal gradients
indicate weak local alignment, while the cause of the mixed-target performance
drop requires evidence about the training trajectory.

\begin{table}[H]
\caption{Final-checkpoint loss and gradient diagnostics (mean $ \pm $ sample SD over three
seeds). Gradient cosine is reported for the final encoder block and the predictor.}
\label{tab:objective_diagnostics}
\centering
\footnotesize
\setlength{\tabcolsep}{3.5pt}
\begin{tabular}{lrrrr}
\toprule
$p_{\mathrm{same}}$ & Cross-view NMSE & Same-view NMSE & Encoder grad. cosine & Predictor grad. cosine \\
\midrule
0 & $\ObjectivePZeroCrossNMSEMean \pm \ObjectivePZeroCrossNMSESD$
& $\ObjectivePZeroSameNMSEMean \pm \ObjectivePZeroSameNMSESD$
& $\ObjectivePZeroEncoderGradientCosineMean \pm \ObjectivePZeroEncoderGradientCosineSD$
& $\ObjectivePZeroPredictorGradientCosineMean \pm \ObjectivePZeroPredictorGradientCosineSD$ \\
.25 & $\ObjectivePTwentyFiveCrossNMSEMean \pm \ObjectivePTwentyFiveCrossNMSESD$
& $\ObjectivePTwentyFiveSameNMSEMean \pm \ObjectivePTwentyFiveSameNMSESD$
& $\ObjectivePTwentyFiveEncoderGradientCosineMean \pm \ObjectivePTwentyFiveEncoderGradientCosineSD$
& $\ObjectivePTwentyFivePredictorGradientCosineMean \pm \ObjectivePTwentyFivePredictorGradientCosineSD$ \\
1 & $\ObjectivePOneCrossNMSEMean \pm \ObjectivePOneCrossNMSESD$
& $\ObjectivePOneSameNMSEMean \pm \ObjectivePOneSameNMSESD$
& $\ObjectivePOneEncoderGradientCosineMean \pm \ObjectivePOneEncoderGradientCosineSD$
& $\ObjectivePOnePredictorGradientCosineMean \pm \ObjectivePOnePredictorGradientCosineSD$ \\
\bottomrule
\end{tabular}
\end{table}

\section{Cardiac Phenotype Prediction and Transfer}
\label{app:idp}

\paragraph{Phenotypes and feature construction.}
The 17 IDPs follow the field definitions in the UK Biobank Data Showcase.
The 13 structural measures comprise LV/RV end-diastolic and end-systolic volumes, LA/RA
maximum and minimum volumes, LV myocardial mass, and four wall-thickness measures. The
four functional measures are average heart rate, LA stroke volume, LA ejection fraction,
and global LV circumferential strain; grouping stroke volume as functional is a reporting
convention. All 17
fields receive equal weight in the overall cardiac-probe score. Tables~\ref{tab:idp_traits_a} and~\ref{tab:idp_traits_b} give the UK Biobank field IDs. Continuous-IDP
probes concatenate the three frozen view representations. Disease classifiers
use elementwise averaging, as specified in Appendix~\ref{app:ukb_disease}.

\subsection{Repeated-CV protocol for ventricular ejection fraction}
\label{app:ef_protocol}
LVEF and RVEF are evaluated separately from the 17-IDP mean and its
13-structural/four-functional groups. Table~\ref{tab:ef} uses C2
($N=15{,}935$) for all seven models. For each model, the frozen 2Ch, 3Ch,
and 4Ch features are concatenated and a supervised standardized linear probe
is fitted on the training participants of each outer fold. The three complete
five-fold partitions are shared across models. Preprocessing fitted from
features and the probe itself use only the corresponding training fold.

\Needspace{10\baselineskip}
For model $m$, let $K_m$ be its reported checkpoint count and let
$\widehat{\boldsymbol y}_{m,j}^{(r)}$ collect the OOF predictions from all five
test folds in repetition $r\in\{1,2,3\}$ for checkpoint $j$. Each participant
contributes one held-out prediction per repetition and checkpoint. For either EF endpoint and each metric
$T\in\{\operatorname{MAE},\operatorname{PearsonR}\}$, the reported aggregation is
\begin{align}
 t_{m,j}^{(r)} &= T\!\left(\boldsymbol y,
                \widehat{\boldsymbol y}_{m,j}^{(r)}\right), &
 a_m^{(r)} &= \frac{1}{K_m}\sum_{j=1}^{K_m}t_{m,j}^{(r)}, \label{eq:ef_checkpoint_aggregation} \\
 \overline a_m &= \frac{1}{3}\sum_{r=1}^{3}a_m^{(r)}, &
 s_m &= \sqrt{\frac{1}{3-1}\sum_{r=1}^{3}
                 \left(a_m^{(r)}-\overline a_m\right)^2}. \label{eq:ef_repetition_summary}
\end{align}
Each metric is computed from pooled OOF predictions within a repetition,
then averaged across checkpoint \emph{metrics} while keeping their predictions
separate. Table~\ref{tab:ef} reports $\overline a_m \pm s_m$, with MAE in EF
percentage points and $R$ the Pearson correlation. The sample SD measures
sensitivity to the three partitions of the fixed cohort, conditional on the
available checkpoints. It quantifies split variability rather than
pretraining-seed variability.

Table~\ref{tab:ef} uses three checkpoints for each of the four internal models
and one for each of the three external systems. The LVEF comparison between
\method{} and Global HEART-JEPA is descriptive. All models share the evaluation
cohort and folds; external pretraining and input adapters differ. C2 excludes
participants in our pretraining manifest, while overlap with external
pretraining manifests remains unverified.

\begin{figure}[!htbp]
\centering
\includegraphics[width=\linewidth]{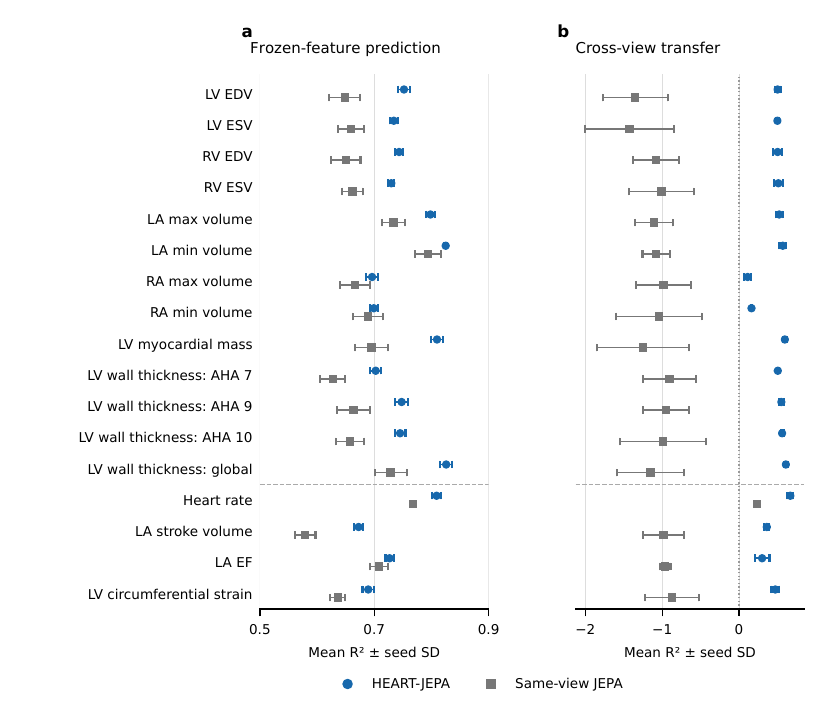}
\caption{Cardiac phenotype prediction and transfer across anatomical views. All 17 verified
fields are shown in anatomical order, with a dashed separator between structural and
functional reporting groups. Points and bars show mean and sample SD across seeds 42--44 on
C2. (a) A standardized OLS probe on concatenated three-view embeddings; five participant
folds averaged within each seed. (b) A source-view standardized ridge probe ($\alpha=1$)
applied to a different view without refitting; five folds and six ordered directions
averaged within each seed. Negative \Rtwo{} indicates worse prediction than the target
test-set mean. Training seeds are the replication unit; traits, folds, and
directions are correlated.}
\label{fig:cardiac_traits}
\end{figure}

\paragraph{Which phenotypes transfer across views?}
Figure~\ref{fig:cardiac_traits} decomposes the matched endpoint comparison on C2. LV global
wall thickness and myocardial mass have transfer scores of
$\TraitHeartTwoFourOneFourZeroTransferMean \pm \TraitHeartTwoFourOneFourZeroTransferSD$ and
$\TraitHeartTwoFourOneZeroFiveTransferMean \pm \TraitHeartTwoFourOneZeroFiveTransferSD$,
followed by LA minimum volume at
$\TraitHeartTwoFourOneOneOneTransferMean \pm \TraitHeartTwoFourOneOneOneTransferSD$, whereas
RA maximum volume reaches
$\TraitHeartTwoFourOneOneFourTransferMean \pm \TraitHeartTwoFourOneOneFourTransferSD$. For RA
maximum and minimum volumes, 4Ch-to-3Ch transfer is negative
($\TraitHeartRAMaxFourToThreeMean \pm \TraitHeartRAMaxFourToThreeSD$ and
$\TraitHeartRAMinFourToThreeMean \pm \TraitHeartRAMinFourToThreeSD$) despite positive
direction-averaged scores. In standard long-axis planning, 3Ch follows the
left ventricular outflow tract rather than the right atrium, whereas 4Ch
includes both atria \citep{kramer2020cmrprotocols}. The 4Ch-to-3Ch RA test
therefore changes both anatomical visibility and feature coordinates. These
factors jointly frame the transfer result; the analysis does not isolate
visibility as the cause, and indirect anatomical correlates can remain
predictive. These exploratory analyses use the common crop prior without
demographic or body-size adjustment.

\paragraph{Post-hoc left- and right-heart structural groups.}
To separate traits with different anatomical coverage, we re-aggregate the
saved C2 per-trait scores into nine LV/LA traits (LV volumes, LV mass, four
wall-thickness measures, and LA volumes) and four RV/RA volume traits.
The groups classify phenotypes by chamber rather than by visibility in an
individual slice. Within each of the three matched
pretraining seeds, we average the already fold- and direction-averaged
transfer \Rtwo{} equally across traits; Table~\ref{tab:anatomical_transfer}
then reports the mean and sample SD across seeds. This summary reuses the
per-trait results without refitting and is reported alongside the original
13-IDP structural mean.

\begin{table}[H]
\caption{Post-hoc anatomical grouping of direct structural transfer on C2.
Values are mean $ \pm $ sample SD across three pretraining seeds after averaging
traits within each group. All six ordered view directions are retained.
These are descriptive summaries of traits with different anatomical coverage.}
\label{tab:anatomical_transfer}
\centering
\small
\begin{tabular}{@{}lcc@{}}
\toprule
Target & LV/LA structural traits ($n=9$) & RV/RA volume traits ($n=4$) \\
\midrule
\method{} & $0.546 \pm 0.021$ & $0.322 \pm 0.020$ \\
Same-view JEPA & $-1.134 \pm 0.395$ & $-1.029 \pm 0.400$ \\
\bottomrule

\end{tabular}
\end{table}

\begin{table}[H]
\caption{Five-fold mean \Rtwo{} for cardiac IDPs and the across-17-IDP mean. \emph{The Mean
column averages all 17 IDPs: the nine shown here and the eight in
Table~\ref{tab:idp_traits_b}.} All four displayed models use the expanded analysis set (17{,}841 participants); RV/LV: right/left ventricle; RA/LA: right/left atrium;
EDV/ESV: end-diastolic/end-systolic volume. Numbers below headers are UK Biobank field IDs.
External models use their respective pretraining and input adapters
(Section~\ref{sec:contextual}).}
\label{tab:idp_traits_a}
\centering
\footnotesize
\setlength{\tabcolsep}{2.5pt}
\begin{tabular}{lrrrrrrrrrr}
\toprule
Model & \shortstack{LVEDV\\24100} & \shortstack{LVESV\\24101}
& \shortstack{RVEDV\\24106} & \shortstack{RVESV\\24107}
& \shortstack{LAmax\\24110} & \shortstack{LAmin\\24111}
& \shortstack{RAmax\\24114} & \shortstack{RAmin\\24115}
& \shortstack{LVmass\\24105} & Mean \\
\midrule
HEART-JEPA & 0.767 & 0.746 & 0.755 & 0.743 & 0.790 & 0.811 & 0.699 & 0.701 & 0.830 & 0.754 \\
CMR Transformer & 0.792 & 0.771 & 0.761 & 0.717 & 0.734 & 0.737 & 0.606 & 0.613 & 0.812 & 0.704 \\
CineMA & 0.805 & 0.716 & 0.780 & 0.739 & 0.833 & 0.791 & 0.786 & 0.748 & 0.840 & 0.689 \\
ViTa-MAE & 0.565 & 0.515 & 0.568 & 0.546 & 0.576 & 0.520 & 0.413 & 0.404 & 0.607 & 0.499 \\
\bottomrule

\end{tabular}
\end{table}

\begin{table}[H]
\caption{Five-fold mean \Rtwo{} for the remaining cardiac IDPs, with the same cohort
as Table~\ref{tab:idp_traits_a} (four displayed models on the expanded analysis set). WT7/9/10/G: LV mean wall thickness in AHA segments
7, 9, 10, and globally; HR: average heart rate; LASV/LAEF: LA stroke volume/ejection
fraction; GCS: global LV circumferential strain. Numbers below headers are field IDs.}
\label{tab:idp_traits_b}
\centering
\footnotesize
\setlength{\tabcolsep}{3.5pt}
\begin{tabular}{lrrrrrrrr}
\toprule
Model & \shortstack{WT7\\24130} & \shortstack{WT9\\24132}
& \shortstack{WT10\\24133} & \shortstack{WTG\\24140}
& \shortstack{HR\\22426} & \shortstack{LASV\\24112}
& \shortstack{LAEF\\24113} & \shortstack{GCS\\24157} \\
\midrule
HEART-JEPA & 0.725 & 0.774 & 0.762 & 0.846 & 0.799 & 0.668 & 0.706 & 0.690 \\
CMR Transformer & 0.618 & 0.683 & 0.689 & 0.752 & 0.785 & 0.645 & 0.598 & 0.650 \\
CineMA & 0.635 & 0.685 & 0.687 & 0.765 & 0.389 & 0.602 & 0.525 & 0.381 \\
ViTa-MAE & 0.433 & 0.471 & 0.492 & 0.550 & 0.603 & 0.439 & 0.347 & 0.433 \\
\bottomrule

\end{tabular}
\end{table}

\section{Directional Retrieval and Available-View Results}
\label{app:missing}

The separate single benchmark checkpoint described in
Section~\ref{sec:contextual} achieves 47.70\%/68.77\% top-1/top-5
retrieval over six ordered view directions (random top-1: 0.0056\%). All scans are
pretraining-disjoint, but 1{,}906 participants have another visit in pretraining; excluding
these participants from both queries and gallery gives 48.74\%/69.67\%, and excluding them only
from queries gives 47.69\%/68.69\%. For comparison, the matched three-seed cross-view
retrieval mean is $\FormalPZeroRetrievalMean \pm \FormalPZeroRetrievalSD\%$.

\begin{table}[H]
\caption{Same-participant retrieval on the expanded analysis set, averaged over six ordered view directions. Queries and gallery each contain 17{,}841 participants. External
models are evaluated as complete pretrained systems with their respective input adapters.}
\label{tab:retrieval}
\centering
\begin{tabular}{lrr}
\toprule
Model & Top-1 & Top-5 \\
\midrule
HEART-JEPA & 47.70 & 68.77 \\
CMR Transformer & 0.50 & 1.74 \\
CineMA & 0.01 & 0.06 \\
ViTa-MAE & 0.02 & 0.07 \\
\bottomrule

\end{tabular}
\end{table}

\begin{table}[H]
\caption{\method{} retrieval by ordered direction on the expanded analysis set (17{,}841 participants).}
\label{tab:directional_retrieval}
\centering
\begin{tabular}{lrr}
\toprule
Direction & Top-1 (\%) & Top-5 (\%) \\
\midrule
2Ch$\rightarrow$3Ch & 43.16 & 65.29 \\
3Ch$\rightarrow$2Ch & 45.87 & 67.46 \\
2Ch$\rightarrow$4Ch & 39.85 & 62.14 \\
4Ch$\rightarrow$2Ch & 41.77 & 63.60 \\
3Ch$\rightarrow$4Ch & 58.72 & 77.45 \\
4Ch$\rightarrow$3Ch & 56.84 & 76.65 \\
\bottomrule
\end{tabular}
\end{table}

\begin{table}[H]
\caption{Available-view mean IDP \Rtwo{} on the expanded analysis set (17{,}841 participants). A separate probe is fitted for each
available-view subset using the same five participant folds. Each score measures
prediction with a probe fitted for that input subset.}
\label{tab:available_views}
\centering
\footnotesize
\begin{tabular}{lrrrr}
\toprule
Available views & \method{} & CMR Transformer & CineMA & ViTa-MAE \\
\midrule
2Ch+3Ch+4Ch & 0.754 & 0.704 & 0.689 & 0.499  \\
3Ch+4Ch & 0.720 & 0.670 & 0.661 & 0.473  \\
2Ch+4Ch & 0.742 & 0.681 & 0.675 & 0.491  \\
2Ch+3Ch & 0.702 & 0.662 & 0.612 & 0.451  \\
2Ch only & 0.655 & 0.584 & 0.535 & 0.389  \\
3Ch only & 0.637 & 0.603 & 0.549 & 0.377  \\
4Ch only & 0.684 & 0.611 & 0.624 & 0.422  \\
\bottomrule

\end{tabular}
\end{table}

\begin{figure}[!htbp]
\centering
\includegraphics[width=\linewidth]{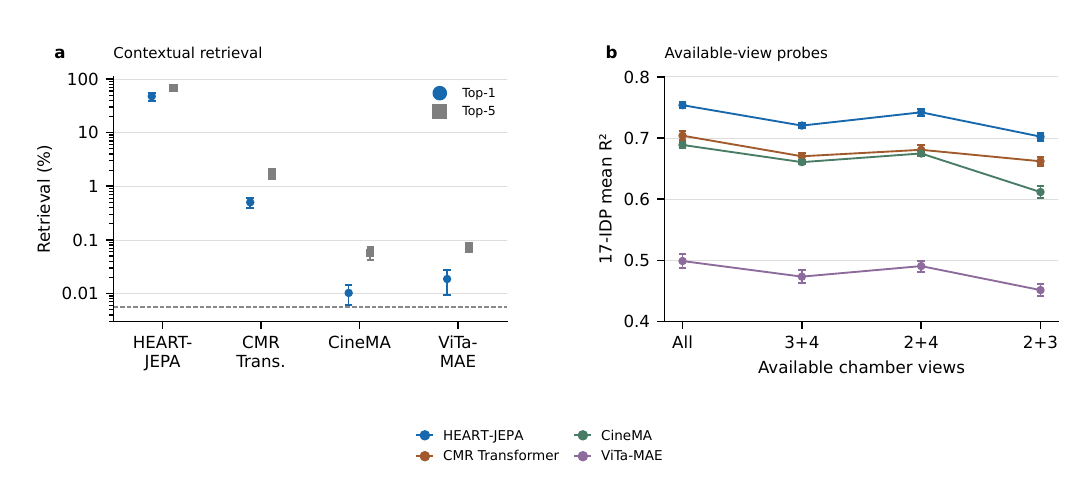}
\caption{Contextual retrieval and available-view cardiac probing on the expanded analysis set (17{,}841 participants). (a) Means and
sample SD across six ordered view directions; the dashed line is
uniform-random top-1. (b) Mean $ \pm $ sample SD across five probe folds, after averaging 17
IDPs within each fold. Each input subset has a separately fitted probe. The
benchmarks compare the supplied frozen representations under their respective
input adapters.}
\label{fig:retrieval_missing}
\end{figure}

\begin{figure}[!htbp]
\centering
\includegraphics[width=\linewidth]{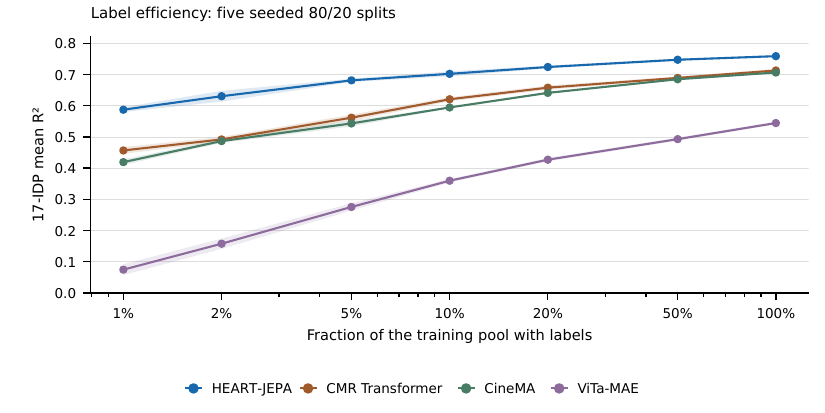}
\caption{Frozen-feature label efficiency for 17 cardiac IDPs. Curves average five seeded
80/20 participant splits (seeds 0--4); bands show the saved population SD across splits
($\mathrm{ddof}=0$), measuring split variability for the evaluated checkpoints.
Fractions refer to the 80\% training pool; standardization and RidgeCV fitting use only
that fraction, and the held-out 20\% is fixed within each seed. This differs from the
five-fold ordinary-linear-regression benchmark; its 100\% endpoint summarizes
the RidgeCV split protocol.}
\label{fig:sample_efficiency}
\end{figure}

\section{External Long-Axis Classification}
\label{app:external}

M\&Ms-2 evaluation uses C5 and frozen 4-chamber long-axis features. Seven
disease-vs-normal classifiers each contrast one disease category against the 75 normal
participants; eight one-vs-rest classifiers include the normal category and use all 360
participants. Both protocols use five stratified folds with split seed 42. A
StandardScaler, up-to-64-component PCA, and class-balanced linear SVM ($C=1$) are fitted
inside each training fold, with no target-dataset encoder fine-tuning. Macro AUC averages
class-specific fold-mean AUCs equally. We report the two protocols separately
because their negative classes differ. These classification benchmarks are
derived from the M\&Ms-2 resource, whose official task is segmentation.
Each model retains its own pretraining (Section~\ref{sec:contextual}).

\begin{table}[H]
\caption{Per-category external classification AUC on M\&Ms-2 (C5) using frozen 4-chamber
features. Values are mean $ \pm $ sample SD across five stratified folds (split seed 42).
$n_+/n_-$ counts positive/negative participants before folding. Bold marks the highest
displayed mean per row, including ties. Macro rows report averages of class
means without pooled error bars.}
\label{tab:external_auc}
\centering
\small
\setlength{\tabcolsep}{3pt}
\begin{tabular}{lcrrrr}
\toprule
\multicolumn{6}{l}{\textbf{(a) Disease vs normal}} \\
Category & $n_+/n_-$ & HEART-JEPA & CMR Transformer & CineMA & ViTa-MAE \\
\midrule
ARR & 35/75 & $0.703 \pm 0.093$ & $\mathbf{0.758} \pm 0.087$ & $\mathbf{0.758} \pm 0.096$ & $0.693 \pm 0.142$ \\
CIA & 35/75 & $0.878 \pm 0.066$ & $\mathbf{0.901} \pm 0.014$ & $0.800 \pm 0.138$ & $0.800 \pm 0.068$ \\
FALL & 35/75 & $0.992 \pm 0.012$ & $\mathbf{1.000} \pm 0.000$ & $0.989 \pm 0.012$ & $0.846 \pm 0.037$ \\
HCM & 60/75 & $\mathbf{0.910} \pm 0.037$ & $0.894 \pm 0.056$ & $0.872 \pm 0.037$ & $0.808 \pm 0.069$ \\
LV & 60/75 & $0.852 \pm 0.031$ & $\mathbf{0.950} \pm 0.065$ & $0.941 \pm 0.023$ & $0.808 \pm 0.068$ \\
RV & 30/75 & $0.696 \pm 0.142$ & $\mathbf{0.818} \pm 0.148$ & $0.713 \pm 0.063$ & $0.629 \pm 0.072$ \\
TRI & 30/75 & $0.907 \pm 0.066$ & $\mathbf{0.967} \pm 0.063$ & $0.878 \pm 0.053$ & $0.798 \pm 0.094$ \\
\midrule
Macro mean & -- & $0.848$ & $\mathbf{0.898}$ & $0.850$ & $0.769$ \\
\midrule
\multicolumn{6}{l}{\textbf{(b) One vs rest}} \\
Category & $n_+/n_-$ & HEART-JEPA & CMR Transformer & CineMA & ViTa-MAE \\
\midrule
ARR & 35/325 & $0.615 \pm 0.120$ & $0.609 \pm 0.166$ & $0.593 \pm 0.145$ & $\mathbf{0.644} \pm 0.094$ \\
CIA & 35/325 & $0.722 \pm 0.080$ & $\mathbf{0.766} \pm 0.109$ & $0.714 \pm 0.095$ & $0.656 \pm 0.146$ \\
FALL & 35/325 & $\mathbf{0.948} \pm 0.043$ & $0.931 \pm 0.035$ & $0.935 \pm 0.033$ & $0.840 \pm 0.104$ \\
HCM & 60/300 & $0.899 \pm 0.050$ & $\mathbf{0.906} \pm 0.018$ & $0.863 \pm 0.069$ & $0.824 \pm 0.076$ \\
LV & 60/300 & $0.728 \pm 0.056$ & $\mathbf{0.863} \pm 0.074$ & $0.836 \pm 0.046$ & $0.624 \pm 0.085$ \\
NOR & 75/285 & $0.752 \pm 0.047$ & $\mathbf{0.900} \pm 0.030$ & $0.777 \pm 0.046$ & $0.658 \pm 0.087$ \\
RV & 30/330 & $0.605 \pm 0.104$ & $0.545 \pm 0.091$ & $\mathbf{0.658} \pm 0.127$ & $0.601 \pm 0.088$ \\
TRI & 30/330 & $0.859 \pm 0.011$ & $\mathbf{0.885} \pm 0.035$ & $0.783 \pm 0.110$ & $0.526 \pm 0.190$ \\
\midrule
Macro mean & -- & $0.766$ & $\mathbf{0.801}$ & $0.770$ & $0.672$ \\
\bottomrule
\end{tabular}

\par\smallskip
\begin{minipage}{\linewidth}
\footnotesize
Dataset category codes: ARR, arrhythmogenic cardiomyopathy; CIA, inter-atrial
communication; FALL, tetralogy of Fallot; HCM, hypertrophic cardiomyopathy; LV/RV, dilated
left/right ventricle; NOR, normal; TRI, tricuspid regurgitation. Fold SD is descriptive:
training sets overlap and categories reuse participants.
\end{minipage}
\end{table}

\section{UKB Disease-Label Probing}
\label{app:ukb_disease}

The \method{}, same-view, and cross-view contrastive checkpoints retain the
shared backbone, pretraining data, and optimization budget; head designs differ
across objective families. All three models use existing frozen features on C2;
this benchmark is distinct from the EF protocol in Appendix~\ref{app:ef_protocol}. We use it as a matched internal-objective comparison rather than a foundation-model leaderboard. External pretrained systems are compared separately in the EF, 17-IDP, available-view, and M\&Ms-2 benchmarks, where their frozen feature exports and input adapters are available.

\paragraph{Participants and endpoints.}
The candidate cohort is C2. We use the earliest available scan with all three long-axis
views for each participant. ICD-10-derived participant lists define five endpoints:
atrial fibrillation (590 cases), atrioventricular block (195), diabetes (650), heart
failure (189), and myocardial infarction (321). Each endpoint is evaluated against the same
14{,}362 participants absent from all seven disease lists. This control
definition is based on recorded labels rather than a clinical assessment of health. Dilated and hypertrophic cardiomyopathy are
excluded from the scored endpoints because of limited case counts but remain exclusion
criteria for controls. Participants can contribute to more than one endpoint.

\paragraph{Frozen probes.}
For each scan we average the 384-dimensional 2Ch, 3Ch, and 4Ch features elementwise and use
the result without PCA or truncation. Within each training fold we fit a StandardScaler
followed by L2-regularized logistic regression ($C=1$, class-balanced weights, LBFGS, at
most 2{,}000 iterations). All models use identical participant-level five-fold stratified
splits for split seeds 42, 43, and 44. Preprocessing and classifier fitting are
restricted to each training fold. \method{}, same-view JEPA, and cross-view contrastive learning each have
three pretrained checkpoints, evaluated on every split.
Elementwise averaging gives a fixed 384-dimensional input rather than the
1{,}152-dimensional concatenation used for continuous IDPs and EF. It imposes
symmetric mixing across views and can favor channel-compatible representations.
The disease results are specific to this aggregation protocol;
mean-versus-concatenation sensitivity remains untested.

\paragraph{Results.}
Table~\ref{tab:ukb_disease} reports ROC-AUC and average precision. The latter
uses the average precision score rather than trapezoidal integration of the
precision--recall curve. Values average folds, then training checkpoints, then split seeds,
and the reported SD is across the three split-level means. \method{} has higher mean
ROC-AUC and AP than same-view JEPA across all five disease endpoints. Against cross-view contrastive
learning it has higher ROC-AUC for atrial fibrillation and heart failure and higher AP on
four endpoints, with contrastive learning ahead on the remaining ROC-AUC endpoints and on
AP for atrial fibrillation. Both cross-view objectives retain information
relevant to the disease labels provided by UK Biobank.

These results quantify discrimination of the available UK Biobank disease
labels. Diagnosis timing relative to MRI and the complete code-selection rules
are absent from the label records, placing prospective risk prediction and
clinical validation outside the scope of this evaluation.

\begin{figure}[!htbp]
\centering
\includegraphics[width=0.85\linewidth]{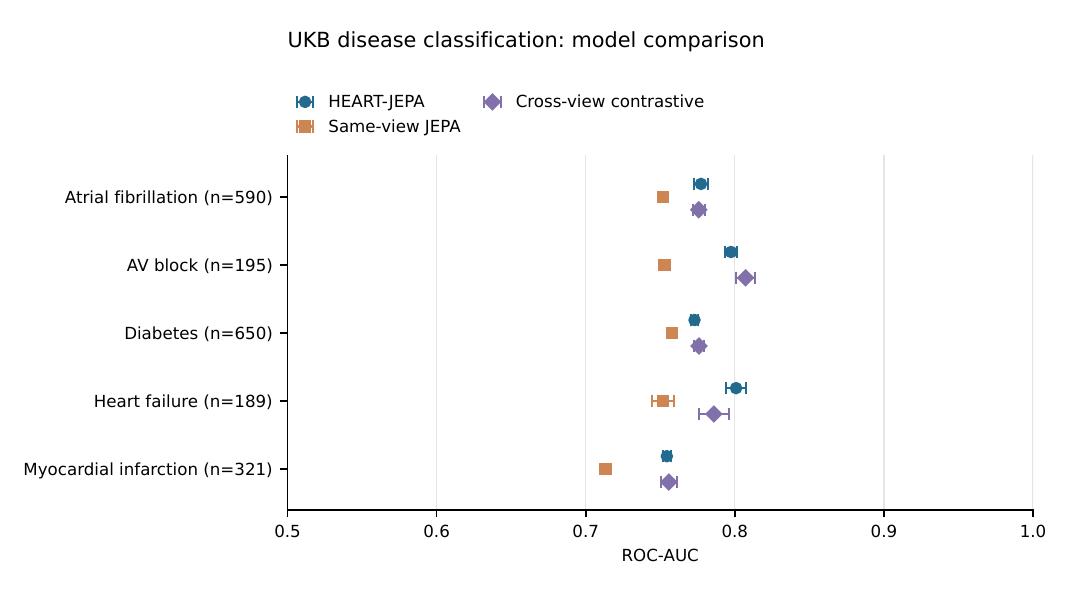}
\caption{UKB disease-label discrimination from frozen three-view representations on C2.
Each endpoint uses the same participants and test folds across models; $n$ denotes cases,
with 14{,}362 controls per endpoint. Points average five folds, three training
checkpoints, and three split seeds. Error bars show SD across split-level means.}
\label{fig:ukb_disease}
\end{figure}

\begin{table}[H]
\centering
\caption{UKB disease-label probing with full 384-dimensional three-view mean features.
Values are mean $ \pm $ SD across three split-level means after averaging folds and training
checkpoints. Same-view denotes the single shared-encoder $p_{\mathrm{same}}=1$
control used across all three views. AP is average precision. The models share
endpoint-specific participants and folds.}
\label{tab:ukb_disease}
\small
\setlength{\tabcolsep}{5pt}
\begin{tabular}{lrrr}
\toprule
Disease ($n_+$) & HEART-JEPA & Same-view & Contrastive \\
\midrule
\multicolumn{4}{l}{\textit{ROC-AUC}} \\
Atrial fibrillation (590) & $0.777 \pm 0.005$ & $0.752 \pm 0.002$ & $0.776 \pm 0.004$ \\
AV block (195) & $0.797 \pm 0.004$ & $0.753 \pm 0.002$ & $0.807 \pm 0.006$ \\
Diabetes (650) & $0.773 \pm 0.003$ & $0.758 \pm 0.002$ & $0.776 \pm 0.003$ \\
Heart failure (189) & $0.801 \pm 0.007$ & $0.752 \pm 0.007$ & $0.786 \pm 0.010$ \\
Myocardial infarction (321) & $0.754 \pm 0.003$ & $0.713 \pm 0.002$ & $0.756 \pm 0.005$ \\
\midrule
\multicolumn{4}{l}{\textit{Average precision (AP)}} \\
Atrial fibrillation (590) & $0.291 \pm 0.006$ & $0.267 \pm 0.001$ & $0.301 \pm 0.002$ \\
AV block (195) & $0.131 \pm 0.003$ & $0.071 \pm 0.002$ & $0.103 \pm 0.001$ \\
Diabetes (650) & $0.176 \pm 0.002$ & $0.151 \pm 0.005$ & $0.172 \pm 0.005$ \\
Heart failure (189) & $0.162 \pm 0.004$ & $0.121 \pm 0.007$ & $0.136 \pm 0.002$ \\
Myocardial infarction (321) & $0.101 \pm 0.004$ & $0.080 \pm 0.002$ & $0.090 \pm 0.006$ \\
\bottomrule
\end{tabular}

\end{table}

\end{document}